\documentclass[journal]{IEEEtran}
\IEEEoverridecommandlockouts
\usepackage[numbers, sort&compress]{natbib}

\usepackage{amsmath,amssymb,amsfonts, bm}
\usepackage[ruled]{algorithm2e}
\usepackage{subfigure}
\usepackage{graphicx}
\usepackage{caption}
\usepackage{textcomp}
\usepackage{xcolor}
\usepackage{authblk}
\usepackage[marginal]{footmisc}
\usepackage{multirow}
\usepackage{multicol}
\usepackage{makecell}
\def\BibTeX{{\rm B\kern-.05em{\sc i\kern-.025em b}\kern-.08em
    T\kern-.1667em\lower.7ex\hbox{E}\kern-.125emX}}

\title{A unified framework based on graph consensus term for multi-view learning}
\author{Xiangzhu Meng\thanks{X. Meng and L. Feng(corresponding author, denoted by $\ast$ ) are with the School of Computer Science and Technology, Dalian University of Technology, Dalian, China(xiangzhu\_meng@mail.dlut.edu.cn; fenglin@dlut.edu.cn)}, Lin Feng\IEEEauthorrefmark{1}, Chonghui Guo\thanks{C. Guo is with the Institute of Systems Engineering, Dalian University of Technology, Dalian, China(dlutguo@dlut.edu.cn)}}

\begin{document}

\maketitle

\begin{abstract}
    In recent years, multi-view learning technologies for various applications have attracted a surge of interest. Due to more compatible and complementary information from multiple views, existing multi-view methods could achieve more promising performance than conventional single-view methods in most situations. However, there are still no sufficient researches on the unified framework in existing multi-view works. Meanwhile, how to efficiently integrate multi-view information is still full of challenges. In this paper, we propose a novel multi-view learning framework, which aims to leverage most existing graph embedding works into a unified formula via introducing the graph consensus term. In particular, our method explores the graph structure in each view independently to preserve the diversity property of graph embedding methods. Meanwhile, we choose heterogeneous graphs to construct the graph consensus term to explore the correlations among multiple views jointly. To this end, the diversity and complementary information among different views could be simultaneously considered. Furthermore, the proposed framework is utilized to implement the multi-view extension of Locality Linear Embedding, named Multi-view Locality Linear Embedding (MvLLE), which could be efficiently solved by applying the alternating optimization strategy. Empirical validations conducted on six benchmark datasets can show the effectiveness of our proposed method.
\end{abstract}

\begin{IEEEkeywords}
Multi-view learning, Unified framework, Graph consensus term, Iterative alternating strategy
\end{IEEEkeywords}

\section{Introduction}
With the rapid development of the information era, more and more data could be obtained from different domains or described from various perspectives, thus multi-view learning technologies\cite{li2018survey, zhao2017multi} have gained extensive attention from researchers in recent years. For examples, an image could be represented by different visual descriptors \cite{ojala2002multiresolution, douze2009evaluation, gao2008image} to reveal its color, texture, and shape information; the document could be translated as different versions via various languages \cite{Amini2009Learning, bisson2012co}; a web-page is usually able to be composed of texts, images, and videos. These different heterogeneous features depict different perspectives to provide complementary information for data description, indicating that each view may contain some knowledge information that other views do not involve. However, classical methods are usually proposed under the single view scenario, which cannot be straightforwardly applied to the multi-view setting. A common solution is to concatenate different views together as one view and then employ single-view algorithms directly for this case. But this concatenation not only lacks physical meaning owing to its specific statistical property in each view, but also ignores the complementary nature of different views. Therefore, the main challenge for multi-view learning is how to effectively combine the information of multiple views and exploit the underlying structures within data.

In recent years, a large amount of multi-view learning approaches have been well investigated in many applications (e.g. classifications \cite{kan2016multi, zhang2019multi, Wang2019A}, clustering \cite{Zheng2018Binary, yang2018multi, wang2019gmc}, etc). Among existing multi-view learning works, one representative category of methods is based on the graph, which is mainly taken into account in this paper. One popular solution \cite{xia2010multiview, Nie2017Auto, Huang2018Self, tian19a} is to consider the weighted combination of different views to explore a common latent space shared by all views in integrating multi-view information. For example, Multiview Spectral Embedding (MSE) \cite{xia2010multiview} was proposed to extend Laplacian Eigenmaps (LE) \cite{belkin2002laplacian} into multi-view setting, which incorporated it with multi-view data to find common low-dimensional representations. Nevertheless, they could not guarantee the complementary effects across different views. For this reason, these algorithms in co-training \cite{Wang2010A, kumar2011co2} and co-regularization \cite{kumar2011co, Niu2019Multi} styles are developed to explore the complementary information among different views. The former iteratively maximizes the mutual agreement on different views to guarantee the consistency of different views. The latter employs co-regularization terms of discriminant functions, added into the objective function, to ensure the consensus among distinct views. Unfortunately, these methods may produce unsatisfactory results when facing such multiple views that are highly related but sightly different from each other. More notably, there aren't still sufficient researches on generalized multi-view frameworks, which are convenient to extend those exiting graph embedding methods based on single view against multi-view tasks, so that the advantages of those single-view works couldn't be fully exploited. What's more, the framework of graph embedding \cite{yan2006graph} implies that most of subspace learning methods \cite{wold1987principal, belhumeur1997eigenfaces, weinberger2006distance} and their kernel extensions \cite{scholkopf1997kernel, torresani2007large, mika1999fisher} could be also cast as special embedding methods based on the graph. Besides, most graph-based deep learning technologies \cite{zhu2020deep, zhu2021graph} have been widely investigated in recent tears.  However, these graph embedding methods cannot be extended into the multi-view setting directly. Therefore, how to extend these works into multi-view setting is the key yet challenging point.

To handle these issues above, we propose a novel model for multi-view learning problems to simultaneously exploit both the diversity and complementary information among different views. Importantly, this model attempts to leverage most existing graph embedding works for single view into a unified formulation. Specifically, to preserve the diversity property of intrinsic information in each view, this model explores the intrinsic graph structure in each view independently; to fully exploit the complementary information among different learned representations, we introduce the graph consensus term, based on heterogeneous graphs, to consider the correlations among multiple views jointly. That is to say, we could utilize the graph consensus term to regularize the dependence among different views and simultaneously obtain the intrinsic structure based on its graph structure or embedding representations for each view. To this end, we formulate the above concerns into a unified framework, named Graph Consensus Multi-view Learning Framework (GCMLF). To facilitate related researches, the proposed framework is utilized to implement the multi-view extension of Locality Linear Embedding \cite{roweis2000nonlinear}, named Multi-view Locality Linear Embedding (MvLLE). Correspondingly, an algorithm based on the alternating direction optimization strategy is provided to efficiently solve MvLLE, which converges to the local optimal value. Finally, extensive experiments based on the applications of document classification, face recognition, and image retrieval validate the ideal performance of our proposed method. In summary, our contributions in this paper could be listed as follows:

\begin{itemize}
\item We propose a novel unified framework multi-view learning problems to leverage most of existing single-view works based on the graph into a unified formula, which utilizes the graph consensus term based on heterogeneous graphs to regularize the dependence among different views.

\item To get the feasible solution of GCMLF, a rough paradigm based on iterative alternating strategy is proposed, which could be verified that it converges to the local optimal value within limited iteration steps.

\item GCMLF is utilized to implement the multi-view extension of Locality Linear Embedding, named Multi-view Locality Linear Embedding (MvLLE), which could be efficiently solved referring to the solving paradigm for GCMLF.
\end{itemize}

The remainder of this paper is organized as follows: in Section \uppercase\expandafter{\romannumeral2}, we briefly review the background of multi-view setting and some methods closely related to our method; in Section \uppercase\expandafter{\romannumeral3}, we describe the construction procedure of our proposed method and its optimization algorithm; in Section \uppercase\expandafter{\romannumeral4}, the proposed framework is utilized to implement the multi-view extension of Locality Linear Embedding; in Section \uppercase\expandafter{\romannumeral5}, extensive experiments on six datasets evaluate the effectiveness of our proposed approach; in Section \uppercase\expandafter{\romannumeral6}, we make the conclusion of this paper.

\section{Related work}\label{related_works}
In this section, we first review a brief comprehension of the related works closed to the proposed method. Then we introduce a multi-view learning method called co-regularized multi-view spectral clustering (Co-reg) \cite{kumar2011co} in detail.

\subsection{Multi-view learning}
Generally, most of multi-view learning methods belong to the category of the graph-based method. Among them, one representative group of multi-view methods \cite{cao2013robust, Nie2017Auto, liu2014multiview} aim to fuse multiple features into single representation, by exploiting the common latent space shared by all views. For example, multi-view sparse coding \cite{cao2013robust, liu2014multiview} combines the shared latent representation for the multi-view information by a series of linear maps as dictionaries. Similarly, Multiple Kernel Learning (MKL) \cite{Lin2011Multiple, Nen2011Multiple, Gu2017Multiple} is also a natural way to integrate different views based on the direct combination of different views, where the work \cite{Lin2011Multiple} learns a common low-dimensional representation with unsupervised or supervised information. However, these methods usually map different views to a common space, which might produce unsatisfactory results because they cannot guarantee the complementarity across different views.

Another typical group of multi-view methods aim to integrate complementary information among different views. Among these works, there are two classes of multi-view methods related to our work, which are based on Canonical Correlation Analysis (CCA) \cite{hardoon2004canonical} and Hilbert-Schmidt Independence Criterion (HSIC) \cite{gretton2005measuring}, respectively. Suppose that two sets of $\bm{X}$ and $\bm{Y}$, consisting of $N$ observations, are drawn jointly from a probability distribution. The former \cite{rupnik2010multi, sharma2012generalized, kan2016multi-view} employs CCA to project the two views into the common subspace by maximizing the cross correlation between two views. It could be expressed as follows:
\begin{equation}
\begin{array}{l}
Corr({\bm{X}},{\bm{Y}}) =  tr\left( {{\bm{W}_X}^T\bm{X}{\bm{Y}}^T{\bm{W}_Y}} \right)\\
\end{array}
\end{equation}
where $\bm{W}_X$ and $\bm{W}_Y$ denote the projecting matrix of the set $\bm{X}$ and the set $\bm{Y}$ respectively. $tr(\cdot)$ is the trace of the matrix. In particular, Multi-View Discriminant Analysis \cite{kan2016multi-view} is proposed to extend LDA \cite{belhumeur1997eigenfaces, mika1999fisher} into a multi-view setting, which projects multi-view features into one discriminative common subspace. Generalized Multiview Analysis (GMA) \cite{sharma2012generalized} solves a joint and relaxed problem of the form of quadratic constrained quadratic program (QCQP) over different feature spaces to obtain a common linear subspace, which generalizes CCA for multi-view scenario, i.e. cross-view classification and retrieval. However, dimensionalities of different views must keep equal with each other in this case. The latter \cite{Niu2014Iterative, Cao2015Diversity, Zhang2016Flexible} explores complementary information by utilizing HSIC to measure the correlations of different views. HSIC measures dependence of the learned representations of different views by mapping variables into a reproducing kernel Hilbert space, which could be expressed as follows:
\begin{equation}\label{HSIC}
\begin{array}{l}
HSIC({\bm{X}},{\bm{Y}}) =  (N-1)^{-2}tr\left(  \bm{K}_X \bm{H} \bm{K}_Y \bm{H} \right)\\
\end{array}
\end{equation}
where $\bm{K}_X$ and $\bm{K}_Y$ denote the Gram matrix of the set $\bm{X}$ and the set $\bm{Y}$ respectively. $\bm{H}=\bm{I}-N^{-1}\bm{1}\bm{1}^T$ centers the Gram matrix $\bm{K}_X$ or $\bm{K}_Y$ to have zero mean in the feature space. Compared to those methods based on CCA, such methods could relieve the restriction of equal dimensionalities for different views. In particular, the work \cite{Niu2014Iterative} employs a kernel dependence measure of HSIC to quantify alternativeness between clustering solutions of two views, which iteratively discovers alternative clusterings. Similarly, the work \cite{Zhang2016Flexible} exploits the complementarity information of multiple views based on HSIC to enhance the correlations (or penalize the disagreement) across different views during the dimensionality reduction, and explores the correlations within each view independently, jointly. However, these works usually incorporate the inner product kernel to construct the HSIC term, which might lead to the issue that we cannot obtain satisfactory performance when facing those nonlinear cases. Differing from those methods above, our proposed graph consensus term cannot only overcome the limitation of dimensional equivalent across views but also fully discover the intrinsic structure information of each view and the complementary information among different views.

\subsection{Co-regularized Multi-view Spectral Clustering}

Co-regularized Multi-view Spectral Clustering (Co-reg) \cite{kumar2011co} aims to propose a spectral clustering framework for multi-view setting. To achieve this goal, Co-reg works with the cross-view assumption that the true underlying clustering should assign corresponding points in each view to the same cluster. For the example of two-view case for the ease of exposition, the cost function for the measure of disagreement between two clusters of the learned embedding $\bm{U}^v$ and $\bm{U}^w$ in the $v$th view and the $w$th view could be defined as follows:
\begin{equation}\label{co_regularizaton1}
    D\left( {{\bm{U}^v},{\bm{U}^w}} \right) =  \left\| \frac{\bm{K}_{\bm{U}^v}}{\left\| \bm{K}_{\bm{U}^v} \right\|_F^2}-\frac{\bm{K}_{\bm{U}^w}}{\left\| \bm{K}_{\bm{U}^w}\right\|_F^2} \right\|_F^2
\end{equation}
where $\bm{K}_{\bm{U}^v}$ is the similarity matrix for the $v$ view and $\left\| \cdot \right\|_F^2$ denotes the Frobenius norm of the matrix. For the convenience of solving the solution, linear kernel is chosen as the similarity measure, that is $\bm{K}_{\bm{U}^v}={\bm{U}^v}{\bm{U}^v}^{^T}$. Substituting this in Eq. (\ref{co_regularizaton1}) and ignoring the constant additive and scaling terms that depend on the number of clusters, the disagreement term $D\left( {{\bm{U}^v},{\bm{U}^w}} \right)$ could be expressed as:
\begin{equation}\label{co_regularizaton2}
    D\left( {{\bm{U}^v},{\bm{U}^w}} \right) =  -tr\left( {{\bm{U}^v}{\bm{U}^v}^{^T}{\bm{U}^w}{\bm{U}^w}^{^T}} \right)
\end{equation}
Co-reg builds on the standard spectral clustering by appealing to the co-regularized framework, which makes the clustering relationships on different views agree with each other. Therefore, combining Eq. (\ref{co_regularizaton2}) with the spectral clustering objectives of all views, we could get the following joint maximization problem for $M$ views:
\begin{equation}
\begin{split}
&\mathop {\min }\limits_{{\bm{U}^1},{\bm{U}^2}, \ldots ,{\bm{U}^M} \in {\mathbb{R}^{N \times k}}} \sum\limits_{v = 1}^m {tr({\bm{U}^v}^{^T}{\bm{L}^v}{\bm{U}^v})}  - \vspace{1cm}\\
&\hspace{2em}\lambda \sum\limits_{1 \le v \neq w \le M} {tr\left( {{\bm{U}^v}{\bm{U}^v}^{^T}{\bm{U}^w}{\bm{U}^w}^{^T}} \right)} \\
&\hspace{3em}s.t.\hspace{1.5em}{\bm{U}^v}^{^T}{\bm{U}^v}{ =  I,} \forall 1 \le v \le M  \\
\end{split}
\end{equation}
where ${\bm{L}^v}$ is the normalized graph Laplacian matrix in the $v$th view and $\lambda$ is a the non-negative hyperparameter to trade-off the spectral clustering objectives and the spectral embedding disagreement terms across different views. In this way, Co-reg implements a spectral clustering framework for multi-view setting. However, choosing linear kernel might lack the ability to capture the nonlinear relationships among different samples in multi-view setting. Besides, there also exists the limitation that the dimensionalities of all views must keep same with each other.

\begin{figure*}[htbp]
\centering
\includegraphics[width=0.9\textwidth]{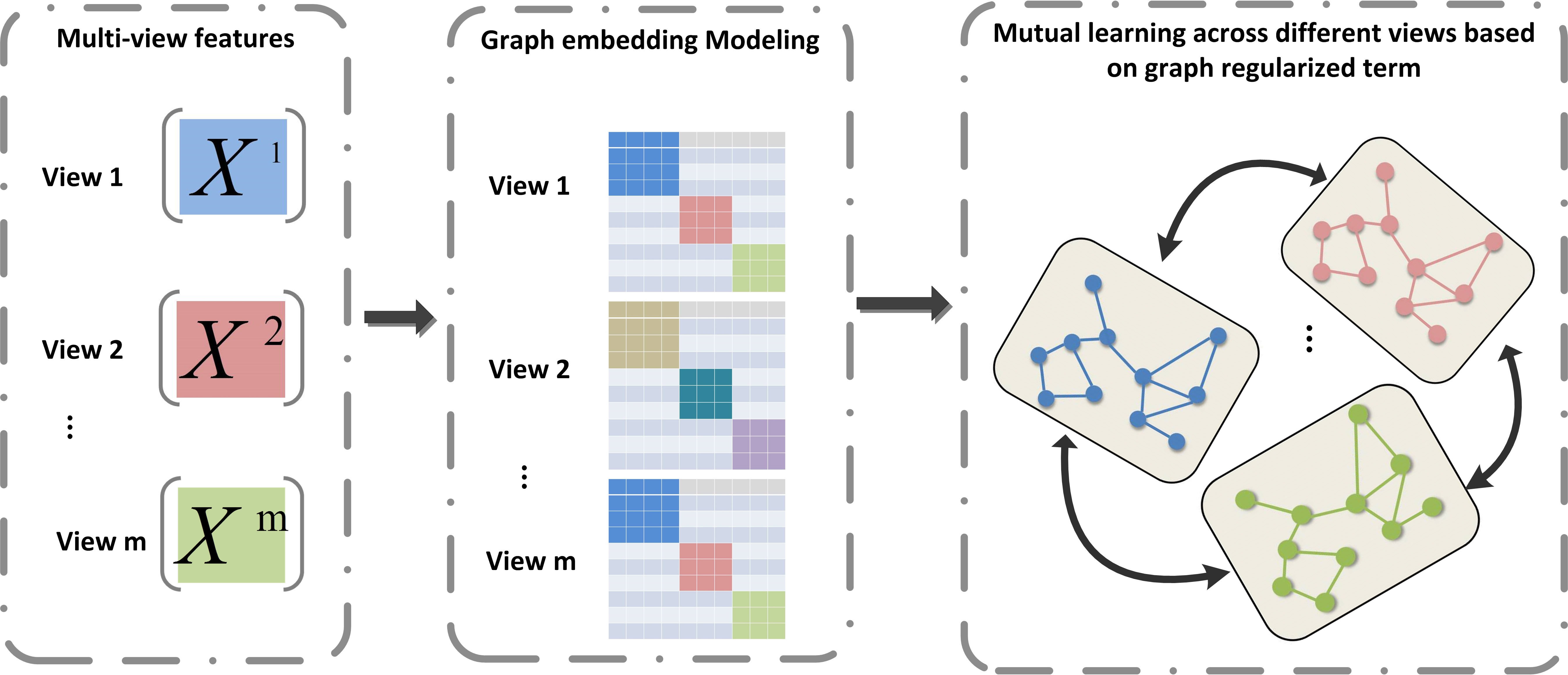}
\caption{The flow chart of the proposed Graph Regularized Multi-view Learning Framework (GCMLF). Given a collection of samples with m views, e.g., $\{\bm{X}^1, \bm{X}^2, \ldots, \bm{X}^m\}$. GCMLF first explores the graph structure in each view by graph embedding model independently, which aims to preserve the diversity property of graph structure information in each view. Then, it utilizes the graph consensus term to regularize the dependence among different views, which makes different views mutually learn. For the example with view $\bm{1}$, we could not only explore the intra-view graph information according to $\bm{X}^1$, but also fully consider inter-view graph structure information more flexibly and robustly. In this way, GCMLF could consider the complementarity among different views and simultaneously obtain the graph embedding for each view.}
\label{flow-chart}
\end{figure*}

\section{Methodology}\label{main_method}
In this section, we discuss the intuition of our proposed framework, named Graph Consensus Multi-view Learning Framework (GCMLF). Here, we propose to introduce the graph consensus term, based on heterogeneous graphs, to regularize the dependence among different views. We first work with two-views to formulate the graph consensus term. Then, the unified multi-view framework is developed for the case of more than two views to enforce multiple views close to each other. For clarity, the flow chart of GCMLF is shown in Fig.\ref{flow-chart}. Correspondingly, a rough paradigm based on iterative alternating strategy is proposed to solve the solution of GCMLF, which could be verified that it converges to the local optimal value. Specifically, we provide one typical case based on two heterogeneous graphs, called Multi-view Locality Linear Embedding (MvLLE). According to the scheme to solve GCMLF, the optimization procedure for MvLLE is presented to complete the case. For convenience, the important notations used in the remainder of this paper are summarized in Table \ref{notations}.

\begin{table}[htbp]
\caption{Important notations used in this paper.}
\label{notations}
\centering
\begin{tabular*}{0.49\textwidth}{@{\extracolsep{\fill}}cl}  
\Xhline{1.2pt}
\textbf{Notation} & \vline \ \textbf{Description}\\
\hline  %
\hline
$\bm{X}^v$ & \vline \ The features set in the $v$th view \\
$\bm{x}_i^v$ & \vline \ The $i$th sample in the $v$th view \\
$\bm{K}^v$ & \vline \ The kernel matrix in the $v$th view \\
$\bm{U}^v$ & \vline \ The embedding in the $v$th view \\
$\bm{G}^v$ & \vline \ The graph matrix defined in $\bm{X}^v$ based on homogeneous graph\\
$\bm{G}_{\ast}^v$ & \vline \ The graph matrix defined in $\bm{U}^v$ based on heterogeneous graph\\
\Xhline{1.2pt}
\end{tabular*}
\end{table}

\subsection{Problem Definition}
Assume that we are given a dataset consisting of $M$ views, the data in the $v$th view ($1 \le v \le M$) could be denoted as $\bm{X}^v = \{\bm{x}_1^v, \bm{x}_2^v, \ldots, \bm{x}_N^v\}$, in which $N$ is the number of samples. The proposed method aims to obtain the graph structure or the embedding in each view under the multi-view setting. We separately employ $\bm{G}^v \in \mathbb{R}^{N \times N}$ and $\bm{U}^v \in \mathbb{R}^{d^v \times N}$ to denote the graph structure or the embedding in the $v$th view, where $d^v$ is the dimensionality of the $v$th view. Differing from the graph  $\bm{G}^v$ defined on $\bm{X}^v$, $\bm{G}_{\ast}^w$ is the graph constructed by the learned embedding $\bm{U}^v$. For the multi-view setting, a naive way is to incorporate all views directly as follows:
\begin{equation}
\begin{split}
&\mathop {\min}\limits_{ \left\{\bm{U}^v \in \mathcal{\bm{C}}^v, 1\le v \le M \right\}}  \sum_{v=1}^M  \mathcal{F}(\bm{G}^v, \bm{U}^v)+\lambda \bm{\Omega}(\bm{U}^v)\\
\end{split}
\end{equation}
where $\mathcal{\bm{C}}^v$ denotes the different constraints on the embedding $\bm{U}^v$. $\mathcal{F}(\cdot, \cdot)$ is the loss function defined on the embedding $\bm{U}^v$ and the graph $\bm{G}^v$, and $\bm{\Omega}(\cdot)$ stands for the smooth regularized term of the embedding $\bm{U}^v$. The positive term $\lambda$ trades-off the loss function $\mathcal{F}(\bm{G}^v, \bm{U}^v)$ and the smooth regularized term $\bm{\Omega}(\bm{U}^v)$. Intuitively, this naive way implements graph embedding problem for each view independently and fails to exploit the diversity information of these multiple views. More importantly, this way neglects the correlations of these multiple views, so that the complementary information among multiple views cannot be made full advantage to enforce all views to learn from each other. Accordingly, how to efficiently discover the complementary information among views is the key point. Besides those works based on CCA or HSIC, traditional solutions usually minimize the difference between the embeddings of pairwise views directly. However, such methods are only suitable for the case that the dimensionalities are equal for different views. For these reasons, it's necessary and worthy to develop a novel co-regularization term with better scalibility and robustness to enforce different views to mutually learn.

\subsection{graph regularization term}
In this paper, we investigate to measure the dependence among all views based on graph structures, which reveals the relationships among all samples in each view. Specifically, we attempt to construct the view-structure consensus in terms of heterogeneous graphs to regularize the dependence between two views. Taking the example with two-view case consisting of the $v$th view and the $w$th view, if two graphs are obtained by the same style of graph approaches, discovering similarly property of individual view, we call such two graphs as homogeneous graphs; in contrast, if two graphs are solved by the different style of graph approaches, we call such two graphs as heterogeneous graphs. When facing the case of homogeneous graphs, directly minimizing the gap between two graphs is to make the relationships among all samples, computed from the $v$th view and the $w$th view, as consistent as possible. However, the diversity information from multiple views might be reduced in this way. For this reason, we introduce the heterogeneous graph consensus term to consider the correlations among multiple views.

For the case of heterogeneous graphs, it's unsuitable to straightforward minimize the semantic gap between the graphs from two views owing to their different construction styles. By design, the graph coefficients could reflect the intrinsic geometric properties of one given view, which are invariant to exactly such transformations. Therefore, we expect their characterization of geometry structure in the one view to be equally valid for the other view on the manifold. That is to say, the relationship between two samples in the $v$th view is expected to be closer if the distance in the $w$th view is larger. Accordingly, we propose the following cost function as measure of dependence between two views:
\begin{equation}\label{heterogeneous_1}
\begin{split}
& Reg(\bm{U}^v, \bm{G}_\ast^w) = \sum\limits_{i,j=1}^{N}{\left\| {\bm{U}_{i}^v-\bm{U}_{j}^v} \right\|_2^2 \bm{G}_{\ast_{ij}}^w}\\
& \quad \quad \quad \quad \quad \quad =tr\left( \bm{U}^v (\bm{D}_\ast^w-\bm{G}_\ast^w) {\bm{U}^v}^T \right) \\
\end{split}
\end{equation}
where $\bm{D}_\ast^w$ denote a diagonal matrix, in which the $i$th diagonal element in $\bm{D}_\ast^w$ is the sum of all elements in the $i$th row of $\bm{G}_\ast^w$.

Besides, when the graph structure specifically reflects the reconstruction relationships among samples, i.e. Low-Rank Representation (LRR) \cite{liu2013robust}, we try to solve the self-representation issue by the following form:
\begin{equation}\label{heterogeneous_2_0}
\begin{split}
& \bm{U}^v = \bm{U}^v\bm{G}_\ast^v+\bm{E}^v\\
\end{split}
\end{equation}
where $\bm{E}^v$ denotes the error term of samples reconstruction. At this time, we investigate to measure the dependence between two views from the aspect of space reconstruction. That is, we expect that reconstruction relationships among samples in the one view could be equally preserved in the other view on the manifold. Therefore, we additionally could utilize the following cost function to measure the consensus between the $v$th view and the $w$th view:
\begin{equation}\label{heterogeneous_2}
\begin{split}
& Reg(\bm{U}^v, \bm{G}_\ast^w) = {\left\| {\bm{U}^v-\bm{U}^v\bm{G}_\ast^w)} \right\|_F^2} \\
& \quad \quad \quad \quad \quad \quad =tr\left( \bm{U}^v (\bm{I}_N - \bm{G}_\ast^w) {({\bm{I}_N} - \bm{G}_\ast^w)}^T  {\bm{U}^v}^T \right) \\
\end{split}
\end{equation}

For convenience, we could further summarize the graph consensus term into a unified form $Reg(\bm{U}^v, \bm{G}_\ast^w)=tr\left( \bm{U}^v \bm{L}^w {\bm{U}^v}^T \right)$ through the Eq.(\ref{heterogeneous_1})-Eq.(\ref{heterogeneous_2}), where $\bm{L}^w$ is just dependent on the graph $\bm{G}_\ast^w$. In above discussion, we provide two formulas of $\bm{L}^w$ based on the consistent preservation between two views. To sum up, we could utilize the graph consensus term $Reg(\bm{U}^v, \bm{G}_\ast^w)$ to co-regularize the dependence among different views and simultaneously obtain the graph structure or embedding for each view.

\subsection{Multi-view learning framework based on graph consensus term}\label{main-framework}

To fully explore the correlations and complementary information among multiple views, we employ the graph consensus term in Eq.(\ref{heterogeneous_1})-Eq.(\ref{heterogeneous_2}) to encourage the new representations of different views to be close to each other. Accordingly, combining graph embedding loss term in each view with graph consensus term among all views, the overall objective function could be formulated as follows:
\begin{equation}\label{total_loss}
\begin{split}
&\mathop {\min}\limits_{ \left\{\bm{U}^v \in \mathcal{\bm{C}}^v, 1\le v \le M \right\}}  \underbrace{\sum_{v=1}^M \left( \mathcal{F}(\bm{G}^v, \bm{U}^v) \right)}_{Graph \ embedding \ loss} + \underbrace{\lambda_{R} \sum_{v=1}^M {\bm{\Omega}(\bm{U}^v)}}_{Normalization \ term}\\
& +  \underbrace{ \lambda_{C} \sum_{v \neq w} {Reg(\bm{U}^v, \bm{G}_{\ast}^w)}}_{Graph \ consensus \ term} \\
\end{split}
\end{equation}
where $\lambda_{R}>0$ and $\lambda_{C}>0$ are two trade-off parameters corresponding to the smooth regularized term and graph consensus term respectively. Under the assumption that space structures in different views could reflect intrinsic properties diversely, the first term ensures that the graphs are constructed for homogeneous structures. The second term guarantees the smoothness within each view independently, and the third term enforces that the learned representations $\left\{\bm{U}^v, 1\le v \le M \right\}$ should learn from each other to minimize the gap between them. In this way, when facing multi-view issues, our proposed framework could deal with the diversity information, smooth regularized terms, and complementary information among multiple views jointly.

\textbf{Optimization procedure:} With the alternating optimization strategy, the Eq.(\ref{total_loss}) could be approximately solved. That is to say, we solve each view at a time while fixing other views. Specifically, with all views but $\bm{U}^v$ fixed, we get the following optimization problem for the $v$th view:
\begin{equation}\label{sub-1}
\begin{split}
&\mathop {\min}\limits_{ \bm{U}^v \in \mathcal{\bm{C}}^v }  \mathcal{F}(\bm{G}^v, \bm{U}^v) + \lambda_{R} \bm{\Omega}(\bm{U}^v)+ \\
&  \lambda_{C} \sum_{1 \le v \neq w}^M { ( Reg(\bm{U}^v, \bm{G}_{\ast}^w)+Reg(\bm{U}^w, \bm{G}_{\ast}^v) )} \\
\end{split}
\end{equation}
Note that in $Reg(\bm{U}^w, \bm{G}_{\ast}^v)$, $\bm{G}_{\ast}^v$ is dependent on the target variable $\bm{U}^v$ and Eq.(\ref{sub-1}) couldn't be directly solved. But if $\bm{G}_{\ast}^v$ is set to be stationary, $Reg(\bm{U}^w, \bm{G}_{\ast}^v)$ will be reduced a constant term on $\bm{U}^v$. Without considering the constant terms, Eq.(\ref{sub-1}) will reduce to the following equation:
\begin{equation}\label{sub-2}
\begin{split}
&\mathop {\min}\limits_{ \bm{U}^v \in \mathcal{\bm{C}}^v }  \mathcal{F}(\bm{G}^v, \bm{U}^v) + \lambda_{R} \bm{\Omega}(\bm{U}^v)+ \lambda_{C} \sum_{1 \le v \neq w}^M {Reg(\bm{U}^v, \bm{G}_{\ast}^w)} \\
\end{split}
\end{equation}
which looks simpler to be solved. Suppose that $\bm{U}^v$ could be effectively calculated by solving the Eq.(\ref{sub-2}), this $\bm{U}^v$ could be continuously used to update $\bm{G}_{\ast}^v$ according to the construction manner of chosen homogeneous graph method, which inspires us to compute $\bm{U}^v$ and $\bm{G}_{\ast}^v$ iteratively.

Hereto, all the variables $\{\bm{U}^v,\bm{G}_{\ast}^v, 1\le v \le M\}$ have been updated completely. The whole procedure to solve Eq.(\ref{total_loss}) is summarized in \textbf{Algorithm \ref{algo-1}}.

\begin{algorithm}
\caption{The optimization for GCMLF}
\label{algo-1}
\LinesNumbered
\KwIn{The multi-view data $\{\bm{X}^v,\forall 1\le v \le M \}$, the hyperparameters $\lambda_{R}$ and $\lambda_{C}$, the loss function $\mathcal{F}(\cdot, \cdot)$, the constraint $\mathcal{\bm{C}}^v$, the homogeneous graph manner for $\bm{G}_{\ast}$.
}

\For{v=1:M}{
    Construct $\bm{G}^v$ in the loss function $\mathcal{F}(\cdot, \cdot)$.

    Initialize $\bm{U}^v$ by minimizing the loss function $\mathcal{F}(\cdot, \cdot)$ under the constraint $\mathcal{\bm{C}}^v$.
}

\While{not converged}{
\For{v=1:M}{
    Update $\bm{G}_{\ast}^v$ for the $v$th view according to the construction manner of the chosen homogeneous graph method.
}
\For{v=1:M}{
    Update $\bm{U}^v$ for the $v$th view by solving Eq.(\ref{sub-2}).
}
}

\KwOut{Learned representations \{$\bm{U}^v, 1\le v \le M$\}.}
\end{algorithm}

\textbf{Convergence analysis:}
Because we adopt the alternating optimization strategy to solve our proposed framework, it's essential to analyze its convergence.

\textbf{Theorem 1.} The objective function in Eq.(\ref{total_loss}) is bounded. The proposed optimization algorithm monotonically decreases the loss value in each step, which makes the solution converge to a local optimum.

\textbf{Proof:} In most cases of graph embedding loss function in $v$th view, $\mathcal{F}(\bm{G}^v, \bm{U}^v)$ is positive. Thus, it's readily to be satisfied that there must exist one view which can make $\mathcal{F}_{min}=\mathcal{F}(\bm{G}^v, \bm{U}^v)>0$ to be smallest among all views. Similarly, we also find that the smooth regularized term $\bm{\Omega}(\bm{U}^v)$ must be greater than 0. For the graph consensus terms among views, we could verify that $tr\left( \bm{U}^v \bm{L}^w {\bm{U}^v}^T \right)$ is positive definite quadratic function if $\bm{L}^w$ is a positive definite matrix. Fortunately, this condition is usually established. Similar to the discussion the loss function in each view, there must exist two closest views which could make $\mathcal{C}_{min}=tr\left( \bm{U}^v \bm{L}^w {\bm{U}^v}^T \right)>0$ to be smallest among all pairwise views. And because the hyperparameters $\lambda_{R}>0$ and $\lambda_{C}>0$, it is provable that the objective value in Eq.(\ref{total_loss}) is greater than $M \mathcal{F}_{min}+M(M-1)\mathcal{C}_{min}$. Therefore, the objective function in Eq.(\ref{total_loss}) has a lower bound.

For each iteration of optimizing problem Eq.(\ref{total_loss}), we could obtain the learned representations \{$\bm{U}^v, 1 \le v \le M$\} by iterative solving the Eq.(\ref{sub-2}), which are corresponding to the exact minimum points of Eq.(\ref{total_loss}) for all views respectively. Under the condition that $\bm{G}_{\ast}^v$ is set to be stationary, the value of the objective function in Eq.(\ref{sub-2}) is non-increasing in each iteration of \textbf{Algorithm \ref{algo-1}}. Thus the alternating optimization procedure will monotonically non-increasing the objective in Eq.(\ref{total_loss}).

Denote the value of loss function in Eq.(\ref{total_loss}) as $\mathcal{H}$, and let ${\{\mathcal{H}^t\}}_{t=1}^T$ be a sequence generated by the iteration steps in \textbf{Algorithm \ref{algo-1}}, where $T$ is the length of this sequence. Based on the above analysis, ${\{\mathcal{H}^t\}}_{t=1}^T$ is a bounded below monotone decreasing sequence. According to the bounded monotone convergence theorem \cite{rudin1964principles} that asserts the convergence of every bounded monotone sequence, the proposed optimization algorithm converges. Accordingly, the \textbf{Theorem 1} has been proved.

\subsection{Discussion with other related methods}
For the proposed graph consensus term, we give a more comprehensive explanation by comparing it with other related methods in this section. Compared with the variants based on CCA, our method is not limited by the dimensional equivalent across different views and more applicable to those nonlinear cases. For the HSIC term in Eq. (\ref{HSIC}), linear kernel is usually used to implement $\bm{K}_X$ and $\bm{K}_Y$. Even though this way is convenient to obtain the optimal solution, the optimization for the nonlinear case is not efficient. Besides, Co-reg might meet the similar issue when facing nonlinear cases. Note that, when the graph consensus term focuses on the similarity among samples in other views, HSIC term and the disagreement term $D\left( {{\bm{U}^v},{\bm{U}^w}} \right)$ in Co-reg could be seen as special cases of the graph consensus term. For example, if $Reg(\bm{U}^v, \bm{G}_{\ast}^w)={\bm{U}^v} \bm{H} {\bm{K}^w} \bm{H} {\bm{U}^v}^{^T}$, it's equivalent to the definition of HSIC term with linear kernel. Differently, we could flexibly choose the common kernel function as similarity measure for $\bm{K}^w$, such as polynomial kernel, Gaussian kernel, etc, which is more applicable for the nonlinear case than HSCI term. Specifically, our proposed method is a more general and robust way to enforce the agreement among different views. In summary, our proposed framework has the following advantages in terms of exploitation for multi-view information and the flexibility of general framework:
\begin{itemize}
\item GCMLF is a unified framework to project multi-view data into ideal subspace for most graph embedding methods, which makes full use of the diversity and complementary information among different views. Differing from those methods minimizing the difference of learned representations among views directly, our proposed framework co-regularizes different views to be close to each other by the graph consensus term based on heterogeneous graphs, meanwhile steadily preserves the intrinsic property of each view on homogeneous graphs.

\item For most of existing multi-view learning frameworks, the limitation of dimensional equivalent makes it not flexible for the extensions of those works. Differing from those methods that only hold under this condition to limit their performance, we could flexibly formulate the dimensionality of each view, which eliminates this limitation. Besides, adopting a suitable graph manner to explore the complementary information among multiple views is beneficial to obtain more robust and promising performance. More importantly, GCMLF could incorporate nonlinear universal cases by exploiting the graph structure information based on learned representations.
\end{itemize}

\section{Specific implement}\label{implement}
In this section, we choose two heterogeneous graph embedding methods, consisting of LE \cite{belkin2003laplacian} and LLE \cite{roweis2000nonlinear}, to provide a typical implement for our proposed framework, named Multi-view Locality Linear Embedding (MvLLE). In fact, LLE and LE are used to construct the graph learning loss term and difference term between two views in Eq.(\ref{total_loss}), respectively.
\subsection{The construction process of MvLLE}
LLE lies on the manifold structure of the samples space to preserve the relationships among samples. Based on the assumption that each sample and its neighbors to lie on or close to a locally linear patch of the manifold, then we obtain the weights matrix $\bm{S}^v \in \mathbb{R}^{N \times N}$ by minimizing the following reconstruction error:
\begin{equation}\label{solve_weights}
    Error \left( \bm{S}^v \right){=}{\sum\limits_{i = 1}^N {\| {\bm{X_i^v} - \sum\limits_{j \in Neighbors\{i\}} {{\bm{S}_{ij}^v}{\bm{X}_j^v}} } \|_2 ^2}}
\end{equation}
where $Neighbors\{i\}$ denotes the neighbors of the $i$th sample $\bm{X}_i^v$. By solving the above equation, we could obtain graph structure $\bm{S}^v$ to reflect intrinsic properties of the samples space. We expect their characterization of local geometry in the original space to be equally valid for local patches on the manifold. Each original sample $\bm{X}_i^v$ is mapped to a new representation. This is done by choosing $d^v$-dimensional coordinates to minimize the following embedding cost function:
\begin{equation}
    Error \left( \bm{U}^v \right){=}{\sum\limits_{i = 1}^N {\| {{\bm{U}_i^v} - \sum\limits_{j \in Neighbors\{i\}} {{\bm{S}_{ij}^v}{\bm{U}_j^v}} } \|_2 ^2}}
\end{equation}
Additionally, we constrain the learned representations $\bm{U}_i^v, 1 \le i \le N$ to have unit covariance. With simple algebraic formulation, the above cost problem can be further transformed as follows:
\begin{equation}\label{lle}
\begin{array}{l}
\mathop {\min }\limits_{\bm{U}^v} \hspace{0.5em}tr(\bm{U}^{v}\bm{{({I-S^v})}^T(I-S^v)}\bm{U^{v^T}})\\
\hspace{0.5em}s.t.\hspace{1em}\bm{U}^{v}\bm{U}^{v^T} = \bm{I}_N
\end{array}
\end{equation}
Hereto, we determine that $\mathcal{F}(\bm{U}^v)$ and $\mathcal{\bm{C}}^v$ are responding to $tr(\bm{U}^{v}\bm{{({I-S^v})}^T(I-S^v)}\bm{U^{v^T}})$ and $\bm{U}^{v}\bm{U}^{v^T} = \bm{I}_N$ respectively.

LE aims at preserving the local neighborhood structure on the data manifold, which constructs the weight matrix that describes the relationships among the samples. Specifically, the similarity matrix $\bm{K}$ is to denote the weight coefficients, which could choose the common kernel function as our similarity measure, such as linear kernel, polynomial kernel, Gaussian kernel and etc. Combining this with the graph consensus term in Eq.(\ref{heterogeneous_1}) between the $v$ view and $w$th view, we could define $\bm{L}^w$ as follows:
\begin{equation}
\begin{split}
& \bm{L}^w = \bm{D}^w-\bm{K}^w\\
\end{split}
\end{equation}
where $\bm{D}^w$ denotes a diagonal matrix and ${\bm{D}_{ii}^w}=\sum\limits_j {{\bm{K}_{ij}^w}}$. By rewriting the normalized matrix $\bm{L}^w$, we could get $\bm{L}^w=\bm{I}_N - {\bm{D}^w}^{ - 1/2}\bm{K}^w{\bm{D}^w}^{ - 1/2}$.

According to the above discussion, we have specified each term in objective function in Eq.(\ref{total_loss}) and its constraint terms. In this way, we could extend single-view based LLE into multi-view setting, named Multi-view Locality Linear Embedding (MvLLE). Based on the above, the whole objective function for MvLLE could be formulated as follows:
\begin{equation}
\begin{split}
& \mathop {\min}\mathcal{\bm{O}}\left( \bm{U}^1, \bm{U}^2, \ldots, \bm{U}^M \right) = \\
& \sum_{v=1}^M tr(\bm{U}^{v}\bm{{({I-S^v})}^T(I-S^v)} \bm{U^{v^T}}) + \lambda_{R} \sum_{v=1}^M \bm{\Omega}(\bm{U}^v)\\
& +  \lambda_{C} \sum_{v \neq w} {tr\left( \bm{U}^v (\bm{I}_N - {\bm{D}^w}^{ - 1/2}\bm{K}^w{\bm{D}^w}^{ - 1/2}) {\bm{U}^v}^T \right)}\\
&\hspace{0.5em}s.t.\hspace{1em}\bm{U}^{v}\bm{U}^{v^T} = \bm{I}_N, 1 \le v \le M \\
\end{split}
\end{equation}
Because the constraint terms normalize the scale of $\{\bm{U}^1, \bm{U}^2, \ldots, \bm{U}^M\}$, the smooth regularized term $\bm{\Omega}(\bm{U}^v)$ could be neglected in the objective function of MvLLE. That is, the above equation could be reduced as follows:
\begin{equation} \label{total_loss_lle}
\begin{split}
& \mathop {\min}\mathcal{\bm{O}}\left( \bm{U}^1, \bm{U}^2, \ldots, \bm{U}^M \right) = \\
& \sum_{v=1}^M tr(\bm{U}^{v}\bm{{({I-S^v})}^T(I-S^v)} \bm{U^{v^T}}) \\
& +  \lambda_{C} \sum_{v \neq w} {tr\left( \bm{U}^v (\bm{I}_N - {\bm{D}^w}^{ - 1/2}\bm{K}^w{\bm{D}^w}^{ - 1/2}) {\bm{U}^v}^T \right)}\\
&\hspace{0.5em}s.t.\hspace{1em}\bm{U}^{v}\bm{U}^{v^T} = \bm{I}_N, 1 \le v \le M \\
\end{split}
\end{equation}

\subsection{Optimization}
Referring to the optimization procedure for GCMLF, the Eq.(\ref{total_loss_lle}) could be approximately solved. When solving the $v$th view, with all views fixed but $\bm{U}^v$, we get the following optimization for the $v$th view:
\begin{equation}
\begin{split}
& \mathop {\min}\mathcal{\bm{O}}\left( \bm{U}^v \right) = tr\left(\bm{U}^{v}\bm{{({I-S^v})}^T(I-S^v)} \bm{U^{v^T}}\right) \\
& +  \lambda_{C} \sum_{1 \le v \neq w}^M {tr\left( \bm{U}^v (\bm{I}_N - {\bm{D}^w}^{ - 1/2}\bm{K}^w{\bm{D}^w}^{ - 1/2}) {\bm{U}^v}^T \right)}\\
&\hspace{0.5em}s.t.\hspace{1em}\bm{U}^{v}\bm{U}^{v^T} = \bm{I}_N \\
\end{split}
\end{equation}

\begin{figure}[htbp]
\centering
\subfigure[Yale dataset]{
\centering
\includegraphics[width=0.22\textwidth]{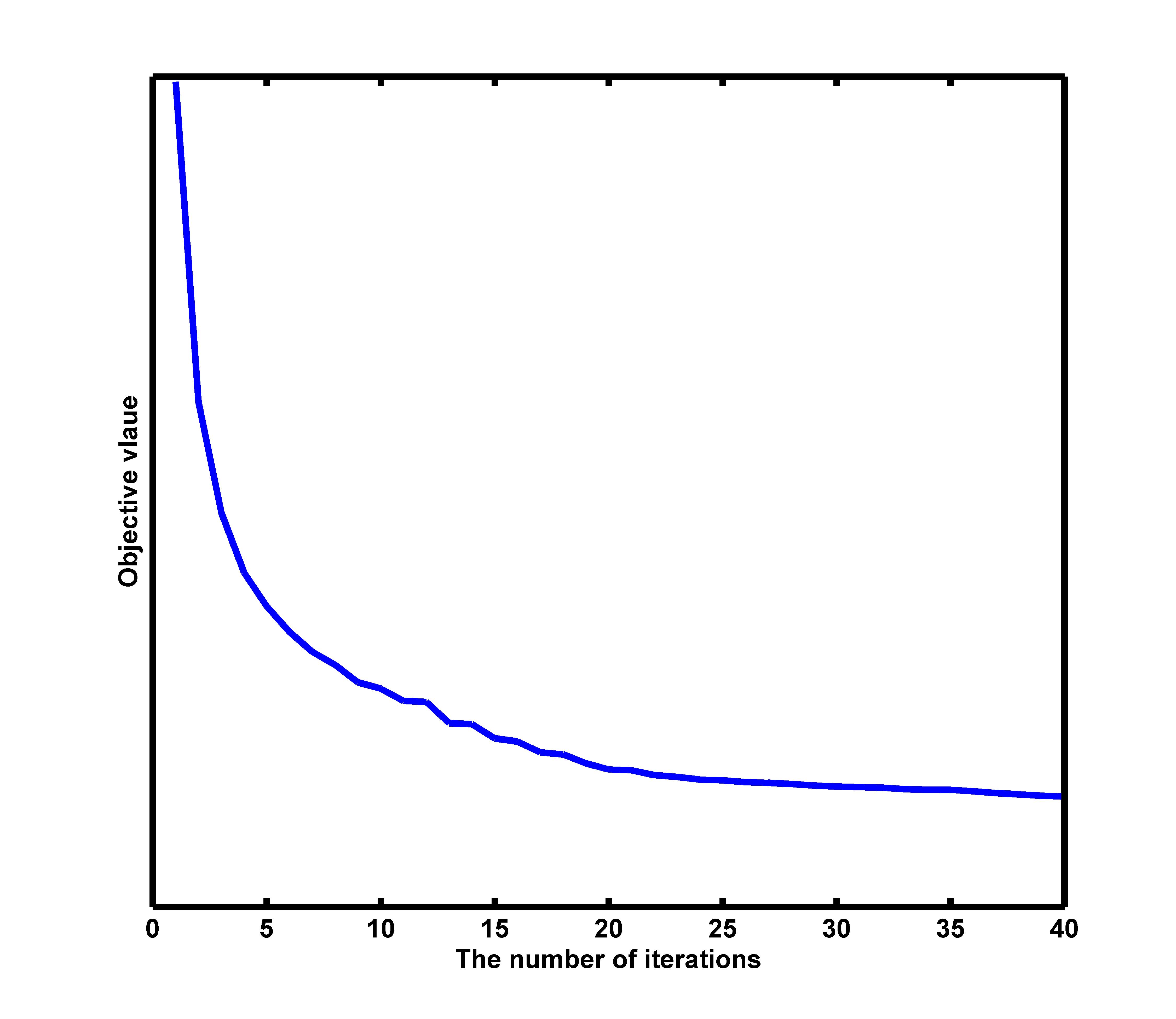}
}
\subfigure[Holidays dataset]{
\centering
\includegraphics[width=0.22\textwidth]{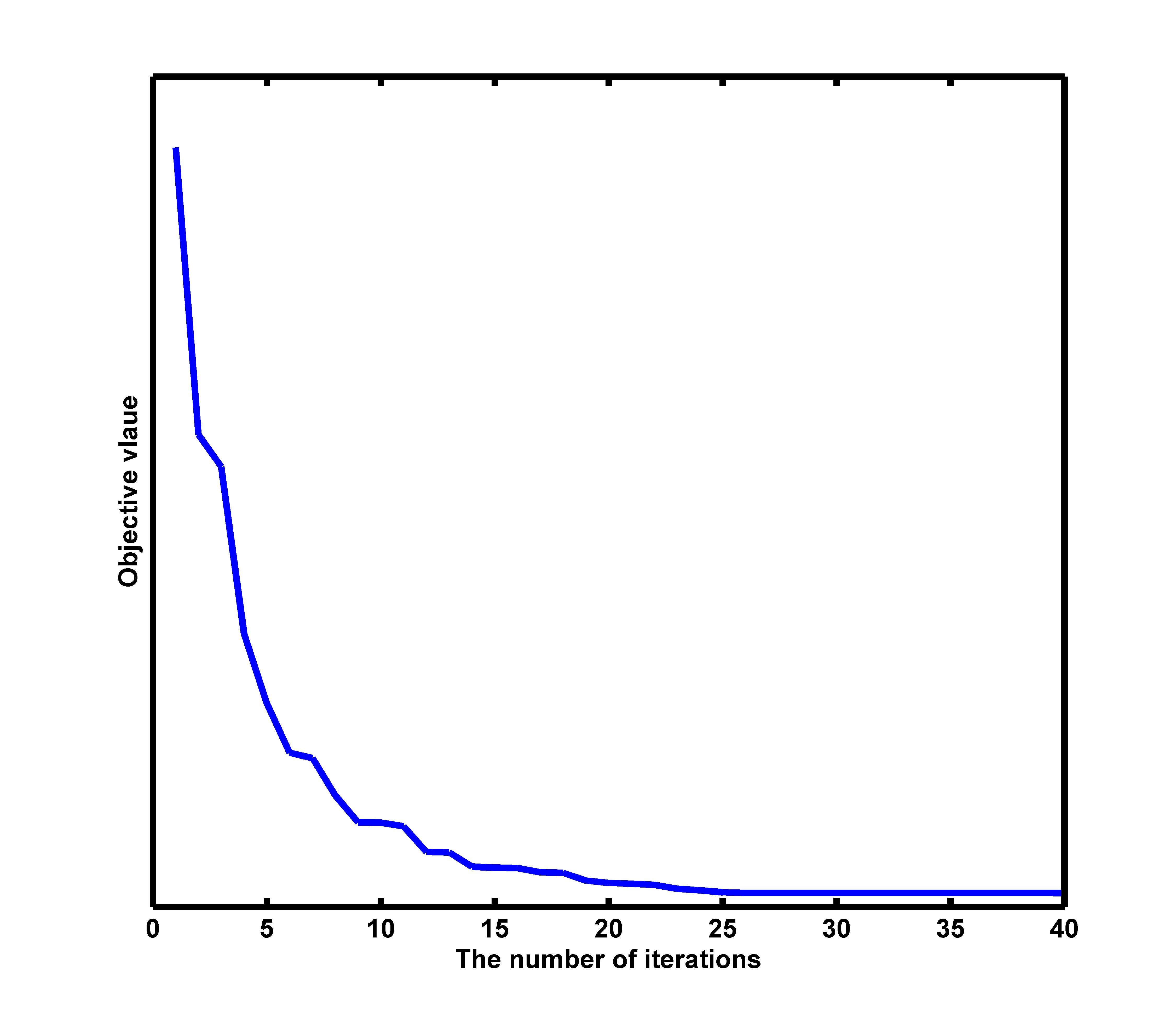}
}
\subfigure[ORL dataset]{
\centering
\includegraphics[width=0.22\textwidth]{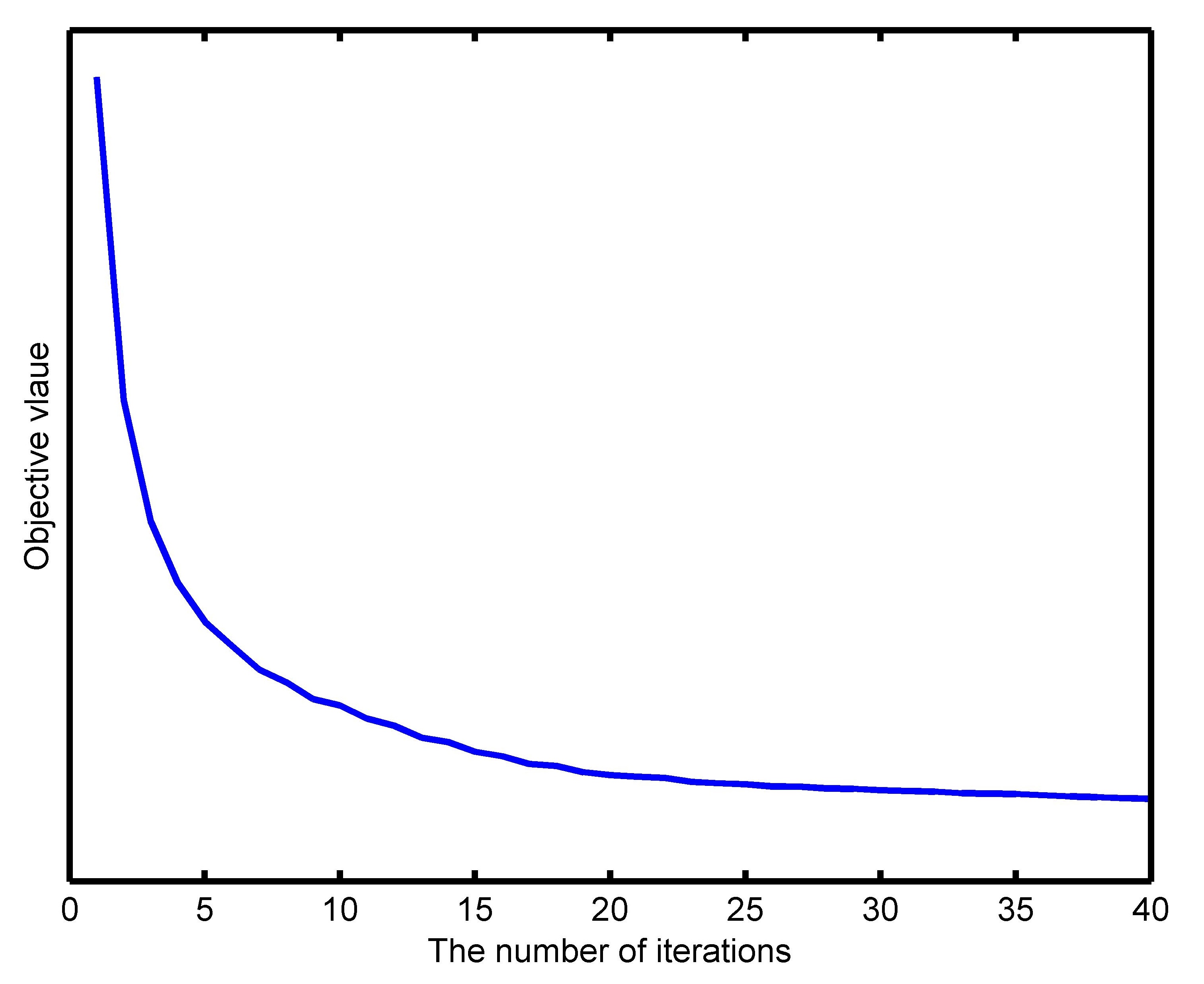}
}
\subfigure[Corel-1K dataset]{
\centering
\includegraphics[width=0.22\textwidth]{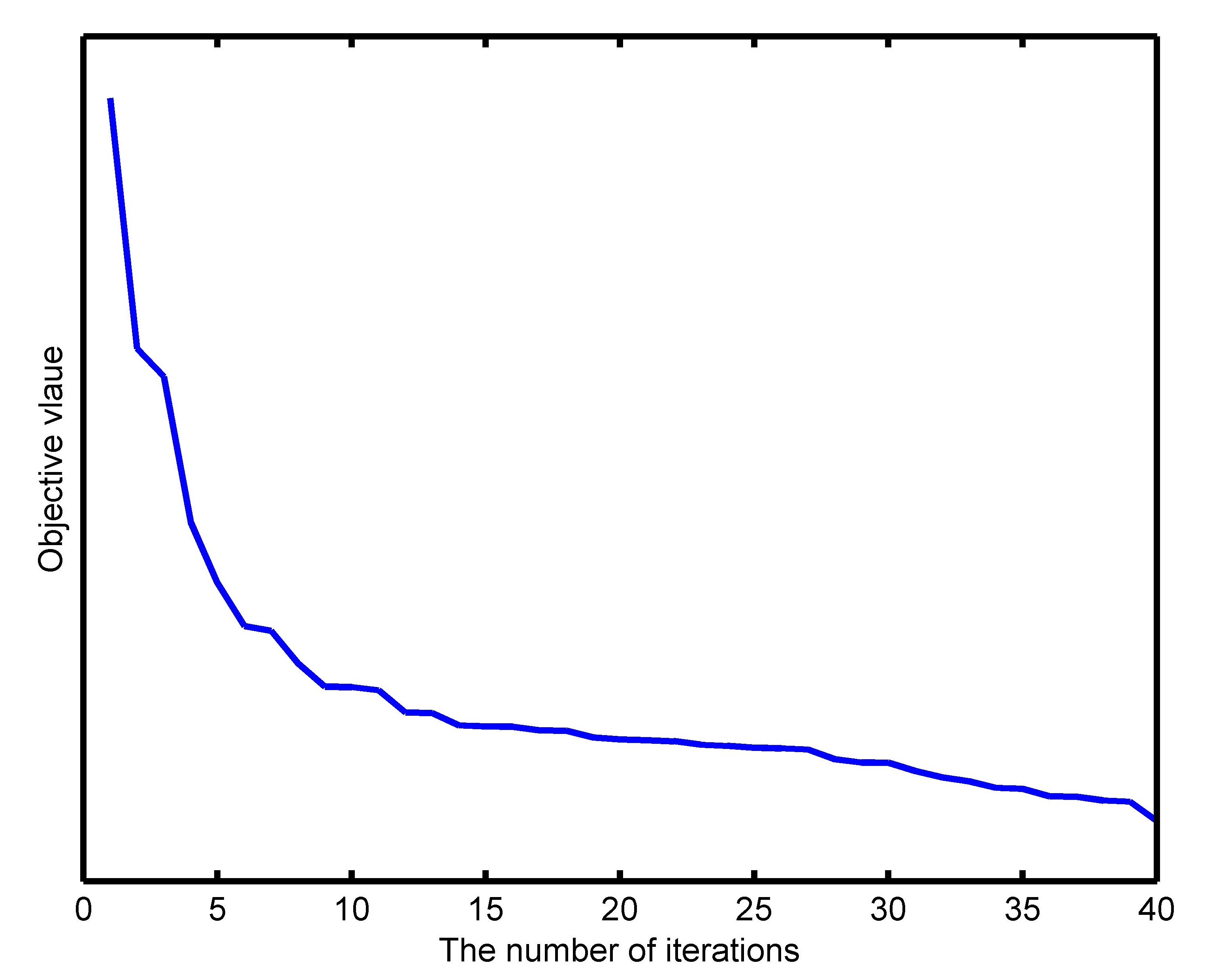}
}

\caption{Convergence validations on four datasets.}
\label{convergence}
\end{figure}

Due to the attributes of the matrix trace, the above equation is equivalent to the following optimization problem:
\begin{equation}\label{lle_view}
\begin{split}
& \mathop {\min}\mathcal{\bm{O}}\left( \bm{U}^v \right) = tr(\bm{U}^{v} ( \bm{{({I-S^v})}^T(I-S^v)} + \\
& \lambda_{C} \sum_{1 \le v \neq w}^M { (\bm{I}_N - {\bm{D}^w}^{ - 1/2}\bm{K}^w{\bm{D}^w}^{ - 1/2}) }  \bm{U^{v^T}}) \\
&\hspace{0.5em}s.t.\hspace{1em}\bm{U}^{v}\bm{U}^{v^T} = \bm{I}_N \\
\end{split}
\end{equation}
Under the constraint condition $\bm{U}^{v}\bm{U}^{v^T} = \bm{I}_N$, the above equation could be efficiently solved by eigenvalue decomposition. In this way, we could solve all the variables $\{\bm{U}^v,\bm{G}_{\ast}^v, 1\le v \le M\}$ iteratively and the whole procedure to solve MvLLE is summarized in \textbf{Algorithm \ref{algo-2}}. According to the convergence analysis for our framework in Section \ref{main-framework}, it could be easily verified that \textbf{Algorithm \ref{algo-2}} for MvLLE will be converged within limited iteration steps. We also use many experiments to verify the convergence property of the proposed method. Fig. \ref{convergence} shows the relation between the objective values and iterations. As shown in Fig. \ref{convergence}, we can see that with the iterations increase, the objective function value of the proposed method decreases fast and reaches a stable point after a few iterations, while the classification accuracy increases dramatically during the first small number of iterations and then reaches the stable high level for these four benchmark databases. For example, for the Holidays dataset, the proposed method reaches the stable point in terms of the classification accuracy within about fifteen iterations. Both theoretical proof and experiments demonstrate that the proposed method can obtain the local optimum quickly and has good convergence property.

\begin{algorithm}
\caption{The optimization procedure for MvLLE}
\label{algo-2}
\LinesNumbered
\KwIn{The multi-view data $\{\bm{X}^v,\forall 1\le v \le M \}$, the hyperparameter $\lambda_{C}$, kernel function $\bm{\kappa}(\cdot, \cdot)$ for similarity matrix $\bm{K}$.
}

\For{v=1:M}{
    Construct $\bm{S}^v$ by solving the Eq.(\ref{solve_weights}).

    Initialize $\bm{U}^v$ by solving the Eq.(\ref{lle}).
}

\While{not converged}{
\For{v=1:M}{
    Update $\bm{K}^v$ for the $v$th view according to kernel function $\bm{\kappa}(\cdot, \cdot)$.
}
\For{v=1:M}{
    Update $\bm{U}^v$ by using eigenvalue decomposition to solve the Eq.(\ref{lle_view}).
}
}

\KwOut{Learned representations \{$\bm{U}^v, 1\le v \le M$\}.}
\end{algorithm}

\subsection{Time complexity}
The computational cost for MvLLE mainly is composed of two parts. One is the construction for the variables $\{\bm{S}^v, i \le v \le M \}$ and the initialization for the variables and $\{\bm{U}^v, i \le v \le M \}$, which solves $\bm{S}^v$ and $\bm{U}^v$ according to Eq.(\ref{solve_weights}) and Eq.(\ref{lle}). The other is to iteratively update $\bm{K}^v$ and $\bm{U}^v$, which needs to perform the computation of similarity matrix and eigenvalue decomposition in each iteration, respectively. It's easy to find that the time complexity of \textbf{Algorithm \ref{algo-2}} is mainly influenced by iteration times and eigenvalue decomposition process. Therefore, its time complexity is about O($T \times M \times N^3$), where $T$ is the iteration times of the alternating optimization procedure. Note that, based on the convergence of \textbf{Algorithm \ref{algo-2}}, the iteration times $T$ will be a limited number.

\subsection{Discussion}
 LLE and LE are two heterogeneous graph embedding methods, in which LLE is used to construct the graph learning loss term and LE is used to regularize the dependence between two views in Eq.(\ref{total_loss}), respectively. Note that, LLE is based on manifold space reconstruction, which aims to preserve reconstruction relationships among samples. Therefore, when LE is utilized to construct the graph learning loss term, we also consider that LLE is used to construct the graph consensus term between two views by Eq.(\ref{heterogeneous_2}). To facilitate the solution, we choose the former to specify the graph learning loss term in Eq. (\ref{total_loss}) in this paper.

\begin{figure*}[htbp]
\centering
\subfigure[Some examples in Yale dataset]{
\centering
\includegraphics[width=0.41\textwidth]{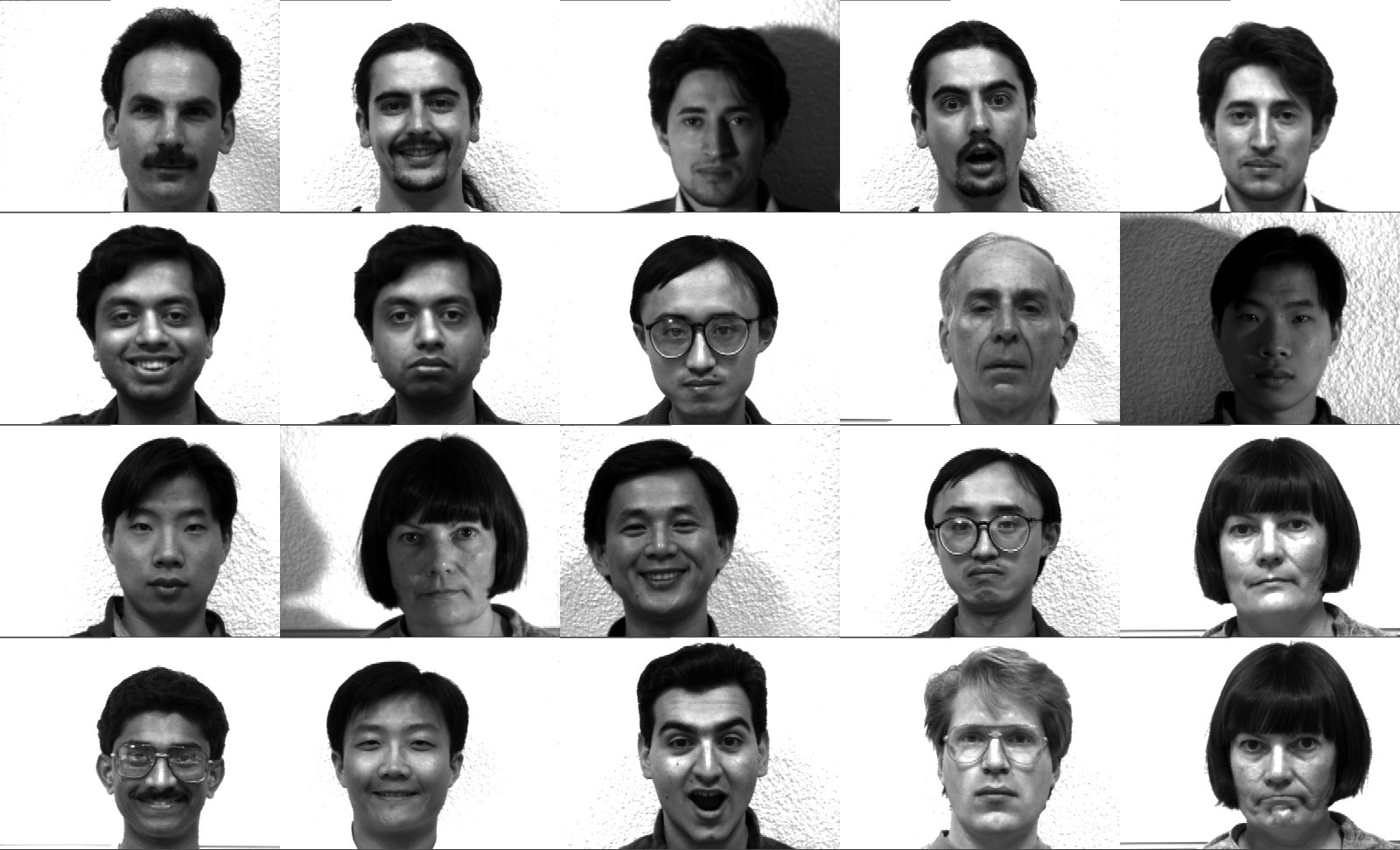}
}
\subfigure[Some examples in Holidays dataset]{
\centering
\includegraphics[width=0.55\textwidth]{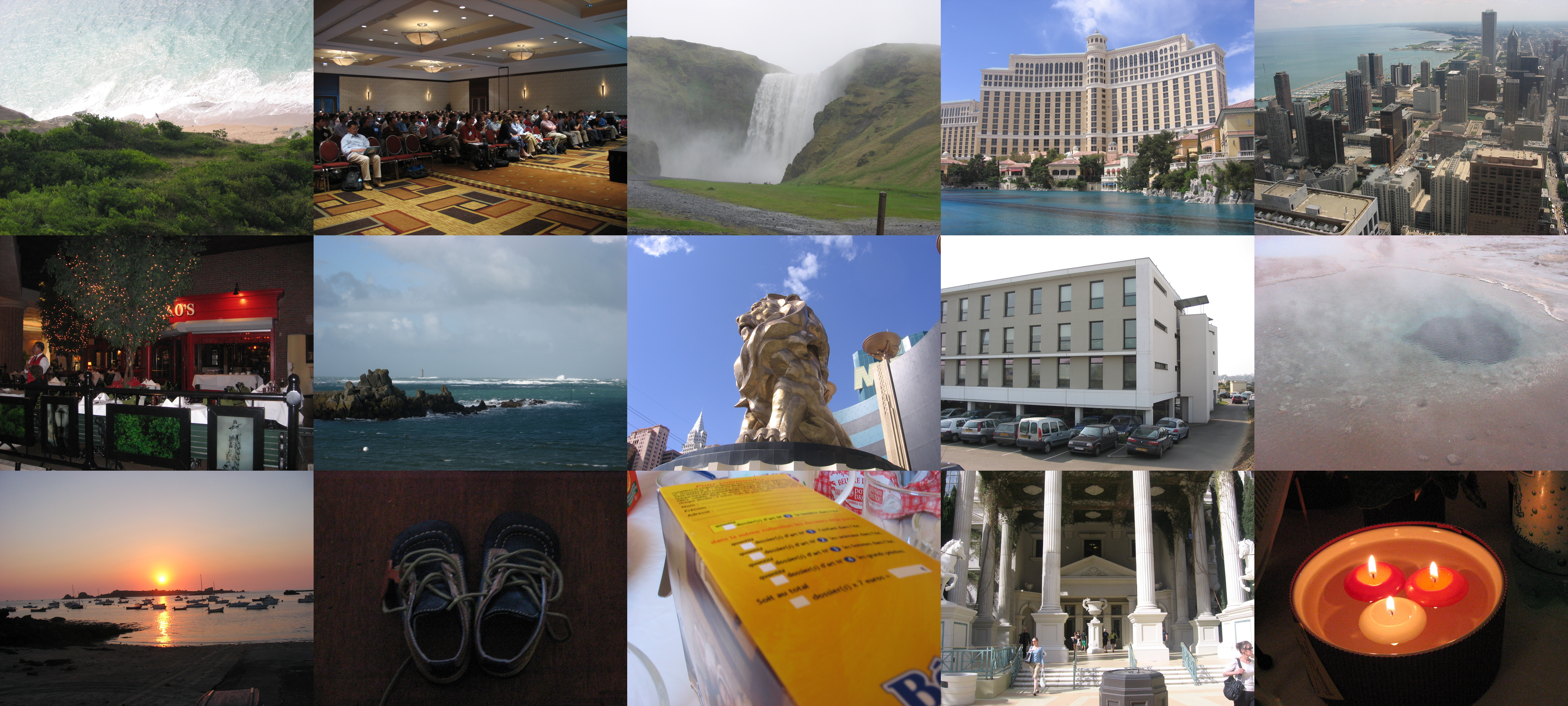}
}
\subfigure[Some examples in ORL dataset]{
\centering
\includegraphics[width=0.34\textwidth]{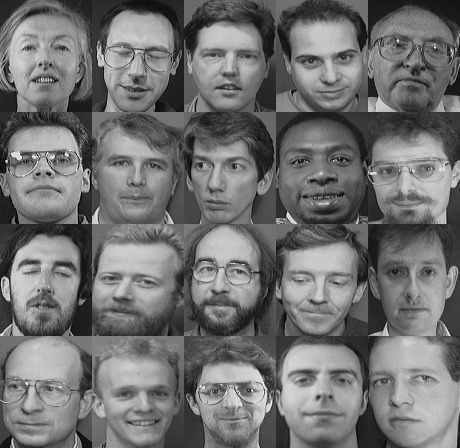}
}
\subfigure[ Some examples in Corel-1K dataset]{
\centering
\includegraphics[width=0.62\textwidth]{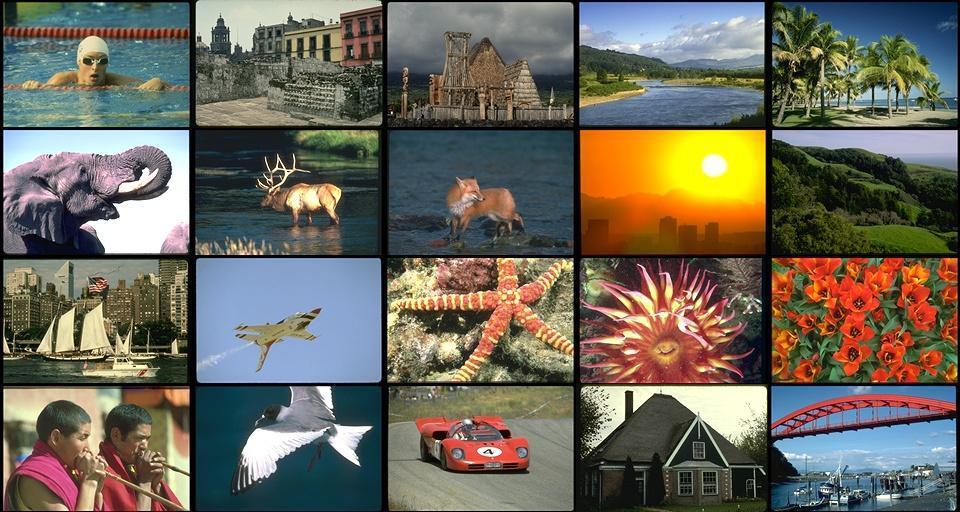}
}

\caption{Examples images in datasets.}
\label{example_images}
\end{figure*}

\section{Experiments}\label{experiments}
In this section, we introduce the details of several experiments on document classification, face recognition, and image retrieval, to verify the effectiveness of our proposed framework. First, six benchmark datasets and related methods for comparison are described detailedly in Section \ref{datasets_and_compared_methods}. Then, we evaluate the performance of our framework by comparing all methods in Section \ref{document}, Section \ref{face}, and Section \ref{image}, respectively. Finally, we take the discussion on the performance of MvLLE based on experimental results on six benchmark datasets in Section \ref{discussion}.

\subsection{Datasets and Compared Methods}\label{datasets_and_compared_methods}
\textbf{Datasets:} In our experiments, six datasets are used to validate the superior performance of our framework, including document datasets (3Source\footnote[1]{http://mlg.ucd.ie/datasets/3sources.html} and Cora\footnote[2]{http://lig-membres.imag.fr/grimal/data.html}), face datasets(ORL\footnote[3]{http://www.uk.research.att.com/facedatabase.html} and Yale\footnote[4]{http://cvc.yale.edu/projects/yalefaces/yalefaces.html}), and image datasets(Corel-1K\footnote[5]{https://sites.google.com/site/dctresearch/Home/content-based-image-retrieval} and Holidays\footnote[6]{http://lear.inrialpes.fr/jegou/data.php}). Two document datasets are two benchmark multi-view datasets. For the face and image datasets, we utilize different descriptors to extract their corresponding multi-view features, in which some samples in these datasets are shown in Fig. \ref{example_images}. The detailed information of these datasets are summarized as follows:
\begin{itemize}
    \item \textbf{3Source} consists of three well-known news organizations: BBC, Reuters, and Guardian, where each news is manually annotated with one of six labels. Because each news source can be used as one view, we choose these news sources as a multi-view benchmark dataset.
    \item \textbf{Cora} contains 2708 scientific publications of seven categories, where each publication document could be described by content and citation. Thus, Cora could be considered as a two-view benchmark dataset.
    \item \textbf{ORL} is collected from 40 distinct subjects, where ten different images are gathered for each subject. For each person, the images are taken at different times, varying the lighting, facial expressions, and facial details.
    \item \textbf{Yale} is composed of 165 faces from 15 peoples, which has been widely used in face recognition. Each person has eleven images, with different facial expressions and facial details.
    \item \textbf{Corel-1K} manually collects one thousand images corresponding to ten categories, such as human beings, buildings, landscapes, buses, dragons, elephants, horses, flowers, mountains, and foods. And there are one hundred images in each category.
    \item \textbf{Holidays} consists of 1491 images corresponding to 500 categories, which are mainly captured for sceneries.
\end{itemize}

To demonstrate the superior performance of our framework, we compare MvLLE with the following methods, where the first two are single-view methods with the most informative view, and the others are multi-view learning methods.
\begin{itemize}
    \item \textbf{BLE} is  Laplacian Eigenmaps (LE) \cite{belkin2003laplacian} with the most informative view, i.e., one that achieves the best performance with LE.
    \item \textbf{BLLE} is Locality Linear Embedding (LLE) \cite{roweis2000nonlinear} with the most informative view, similar to BLE.
    \item \textbf{MSE} \cite{xia2010multiview} is a multi-view spectral method based on global coordinate alignment.
    \item \textbf{CCA} \cite{hardoon2004canonical} is used to deal with multi-view problems by maximizing the cross correlation between two views.
    \item \textbf{Co-reg} \cite{kumar2011co} is a multi-view spectral embedding by regularizing different views to be close to each other.
    \item \textbf{AMGL} \cite{Nie2017Auto} is an auto-weighted multiple graph learning method, which could allocate ideal weight for each view automatically.
\end{itemize}

\subsection{Document Classification}\label{document}
In this section, we evaluate the experimental results of the document classification tasks on 3Source and Cora datasets. For these two datasets, we randomly select 50\% of the samples as training samples and the remaining 50\% of the dataset as testing samples every time. All the methods are conducted to project all samples to the same dimensionality. Specifically, the dimensions of the embedding obtained by all methods all maintain 20 and 30 dimensions. We adopt 1NN as the classifier to classify the testing ones. After conducting this experiment 30 times with different random training samples and testing samples, we calculate the mean classification accuracy (MEAN) and max classification accuracy (MAX) on 3Source and Cora datasets as the evaluation index for all methods. Then, we can summary the evaluation indexes of MEAN and MAX results in Table \ref{3Source} and Table \ref{Cora}.

\begin{table}[!htb]
\caption{The classification accuracy on 3Source dataset.}
\label{3Source}
\centering  %
\begin{tabular}{lllll}  
\hline
Methods & \multicolumn{2}{c}{Dims=20} & \multicolumn{2}{c}{Dims=30} \\
 & MEAN(\%) & MAX(\%) & MEAN(\%) & MAX(\%)\\
\hline
BLE & 66.47 & 74.11 & 59.72 & 69.41 \\
BLLE & 66.50 & 76.71 & 66.78 & 75.94 \\
MSE & 50.47 & 57.64 & 46.86 & 60.00 \\
Co-reg  & 81.25 & 87.05 & 78.50 & 85.88\\
CCA & 53.88 & 76.45 & 54.37 & 73.56 \\
AMGL & 49.92 & 57.64 & 48.15 & 56.47 \\
MvLLE & \textbf{82.64} & \textbf{89.41} & \textbf{79.70} & \textbf{90.9} \\
\hline
\end{tabular}
\end{table}

\begin{table}[!htb]
\caption{The classification accuracy on Cora dataset.}
\label{Cora}
\centering  %
\begin{tabular}{lllll}  
\hline
Methods & \multicolumn{2}{c}{Dims=20} & \multicolumn{2}{c}{Dims=30} \\
 & MEAN(\%) & MAX(\%) & MEAN(\%) & MAX(\%)\\
\hline
BLE & 58.98 & 60.85 & 61.05 & 63.44 \\
BLLE & 59.84 & 63.61 & 60.86 & 65.31 \\
MSE & 64.65 & 66.24 & 67.72 & 69.64 \\
Co-reg & 55.73 & 57.45 & 57.19 & 59.01 \\
CCA & 71.11 & 72.35 & 71.52 & 72.05 \\
AMGL & 63.71 & 65.73 & 66.90 & 69.57 \\
MvLLE & \textbf{73.7} & \textbf{75.23} & \textbf{73.45} & \textbf{75.84} \\
\hline
\end{tabular}
\end{table}

Through the experimental results of Tables \ref{3Source}-\ref{Cora}, it's clear that the proposed MvLLE is significantly superior to its counterparts in most situations. Among the comparing methods, CCA is close to the proposed MvLLE on classification performance, which might take more advantages of complementary information than other compared methods on 3Source and Cora datasets. Compared with other multi-view methods, the performance of our MvLLE is more stable. For example, Co-reg achieves promising results on 3Source dataset while the performance degrades sharply on the Cora dataset.

\subsection{Face Recognition}\label{face}

\begin{figure}
\centering
\includegraphics[width=0.48\textwidth]{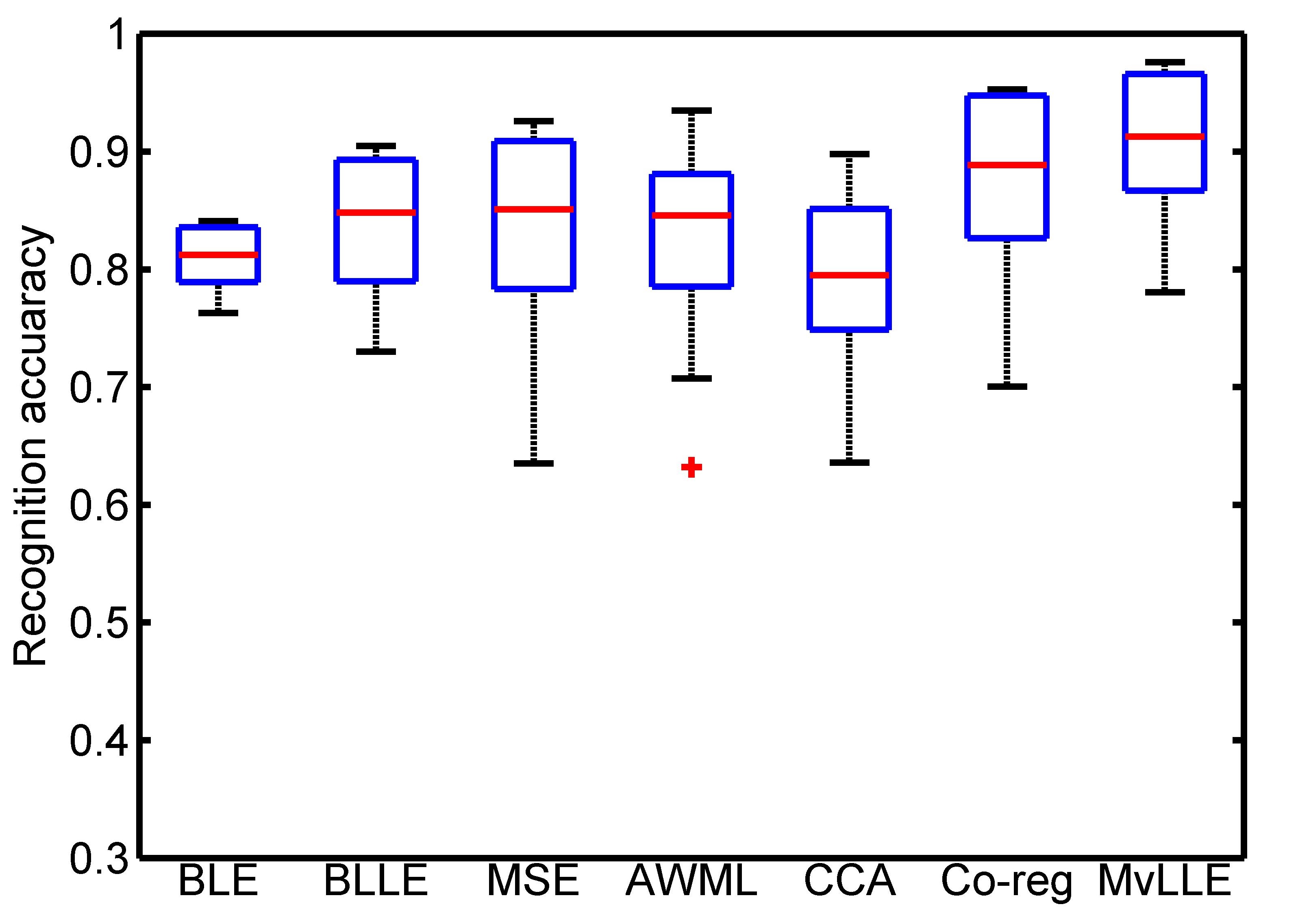}
\caption{The face recognition accuracy on Yale dataset.}
\label{yale_result}
\end{figure}

\begin{figure}
\centering
\includegraphics[width=0.48\textwidth]{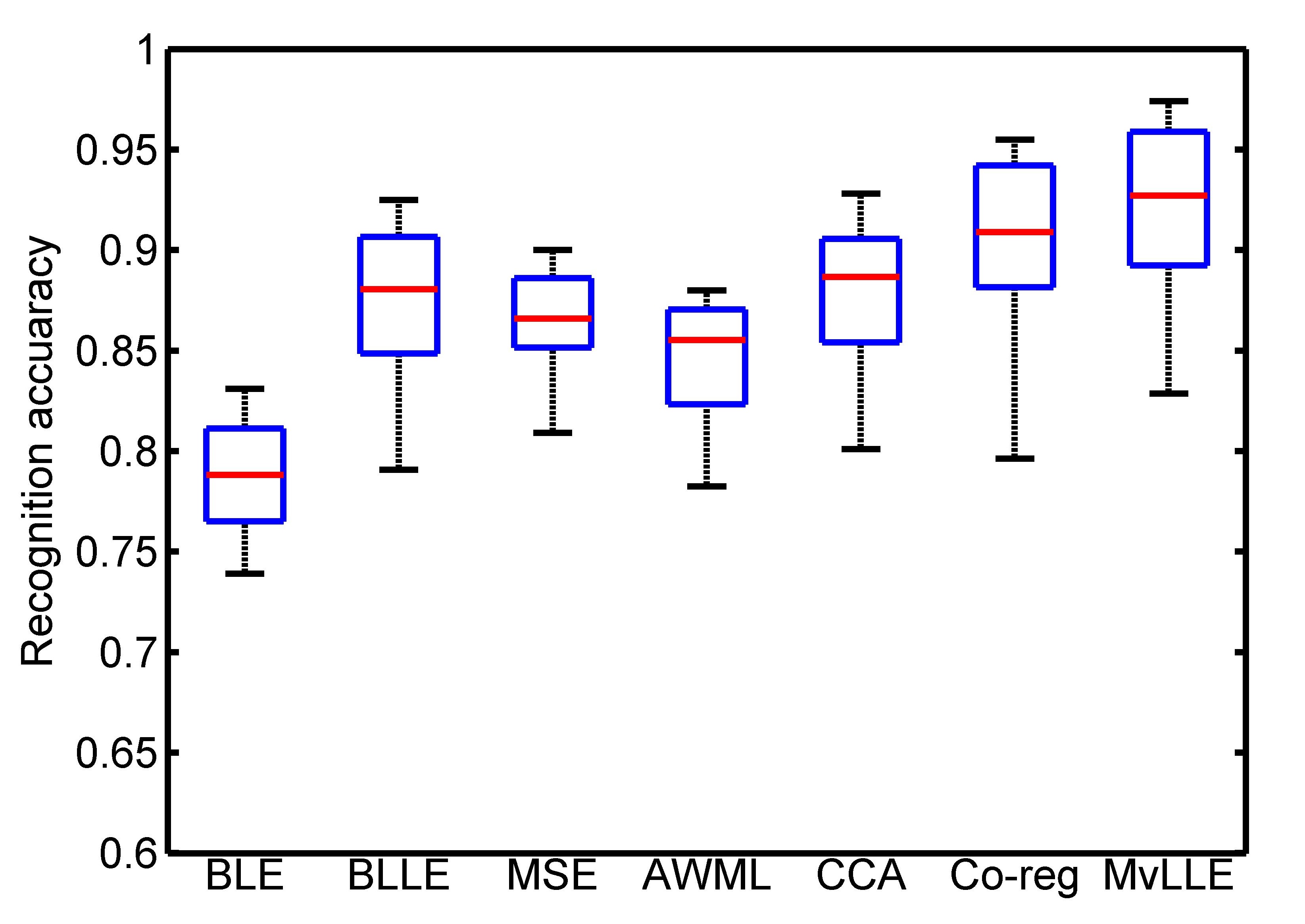}
\caption{The face recognition accuracy on ORL dataset.}
\label{orl_result}
\end{figure}

In this section, we evaluate the experimental results of the face recognition tasks on Yale and ORL datasets. For these two datasets, we first extract their multi-view features by the different image descriptors including EDH \cite{gao2008image}, LBP \cite{ojala2002multiresolution} and Gist \cite{douze2009evaluation}. Then, all the methods are conducted to project all samples to the same dimensionality and the 1NN classifier is adopted to calculate the recognition results, where the dimension of the embedding obtained by all methods all maintains 30 dimensions. Note that we randomly select 50\% of the samples as training samples and the remaining 50\% of the samples as testing samples every time and run all methods 30 times with different random training samples. Because the task of face recognition mainly cares about the recognition accuracy, we choose the recognition accuracy as the evaluation index in this part. The boxplot figures of accuracy values of all methods on Yale and ORL datasets are shown in Fig. \ref{yale_result}and Fig. \ref{orl_result}.

Through the experiment results of the above two experiments in Figs. \ref{yale_result}-\ref{orl_result}, the multiple view performances are usually better than the independent view. This demonstrates that multiple views can improve the performance of face recognition. Among these multi-view methods, we can find that MvLLE outperforms its comparing methods in most situations, which shows the superiority of the proposed framework. Besides our MvLLE, Co-reg obtains stably better than other methods on the performance of face recognition, which takes more advantages of complementary information than other comparing methods on Yale and ORL face datasets.

\subsection{Image Retrieval}\label{image}
\begin{figure*}[htbp]
\centering
\subfigure[Precision]{
\centering
\includegraphics[width=0.45\textwidth]{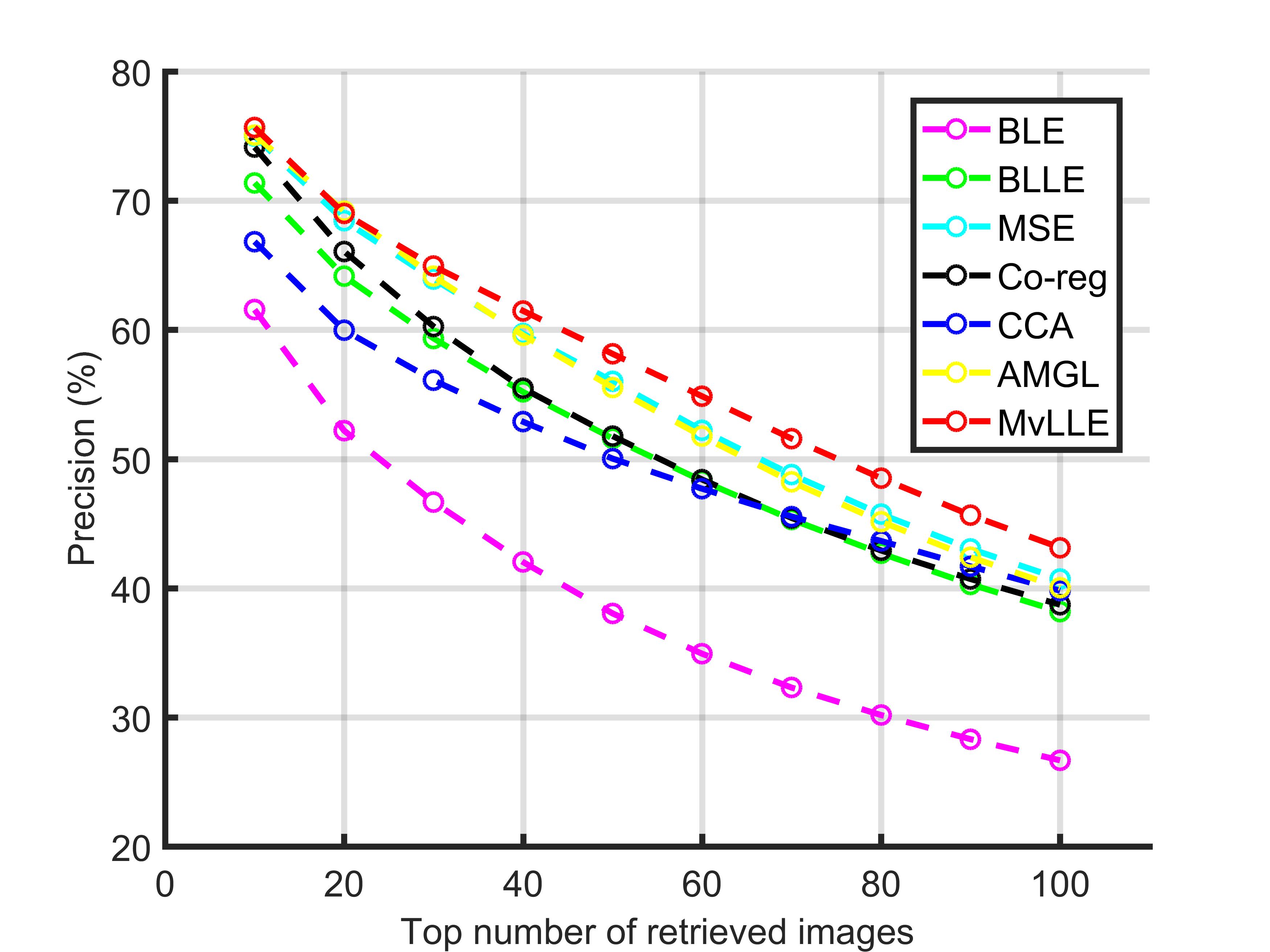}
}
\centering
\subfigure[Recall]{
\centering
\includegraphics[width=0.45\textwidth]{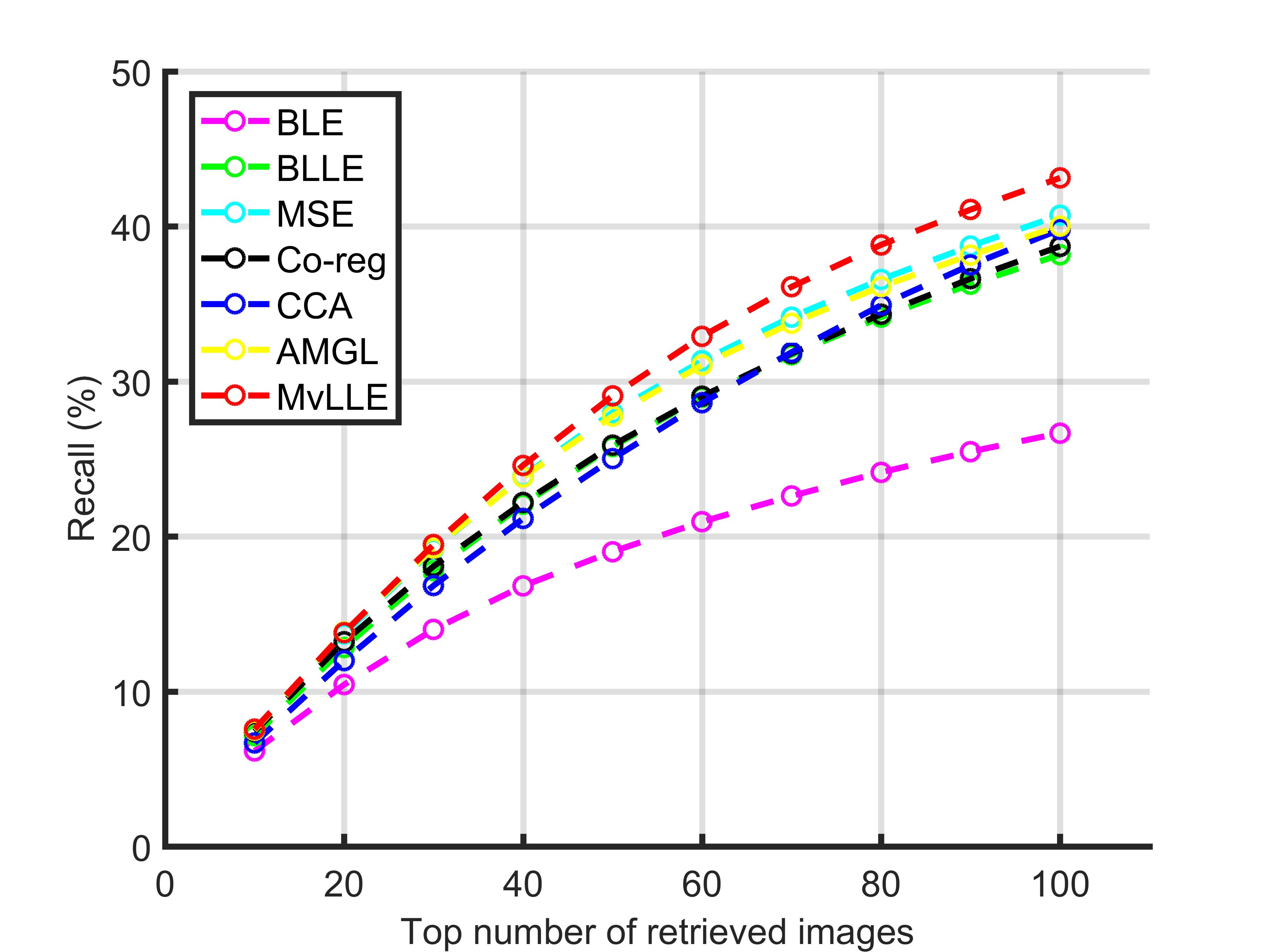}
}

\subfigure[PR-Curve]{
\centering
\includegraphics[width=0.45\textwidth]{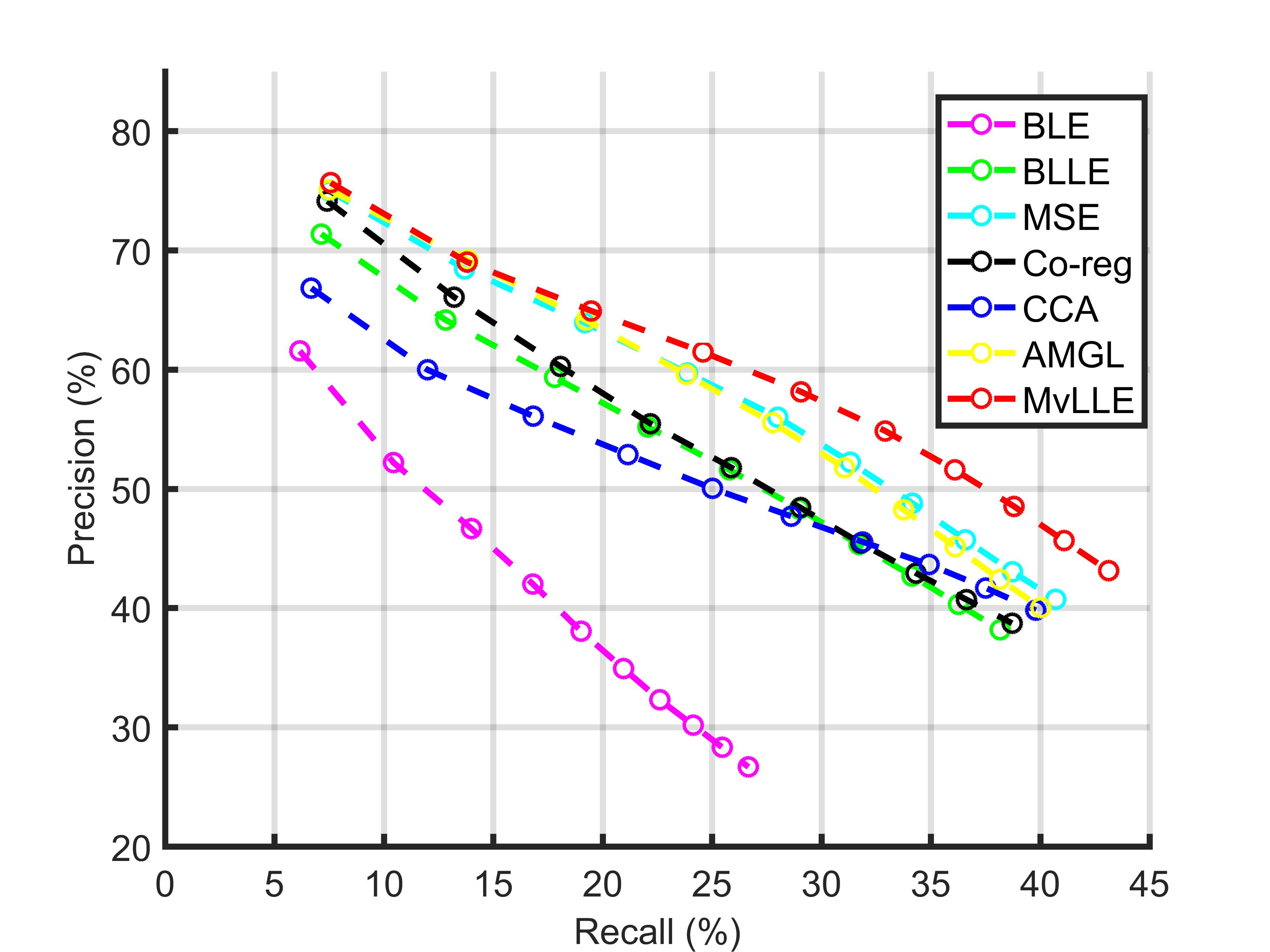}
}
\subfigure[$F_1$-Measure]{
\centering
\includegraphics[width=0.45\textwidth]{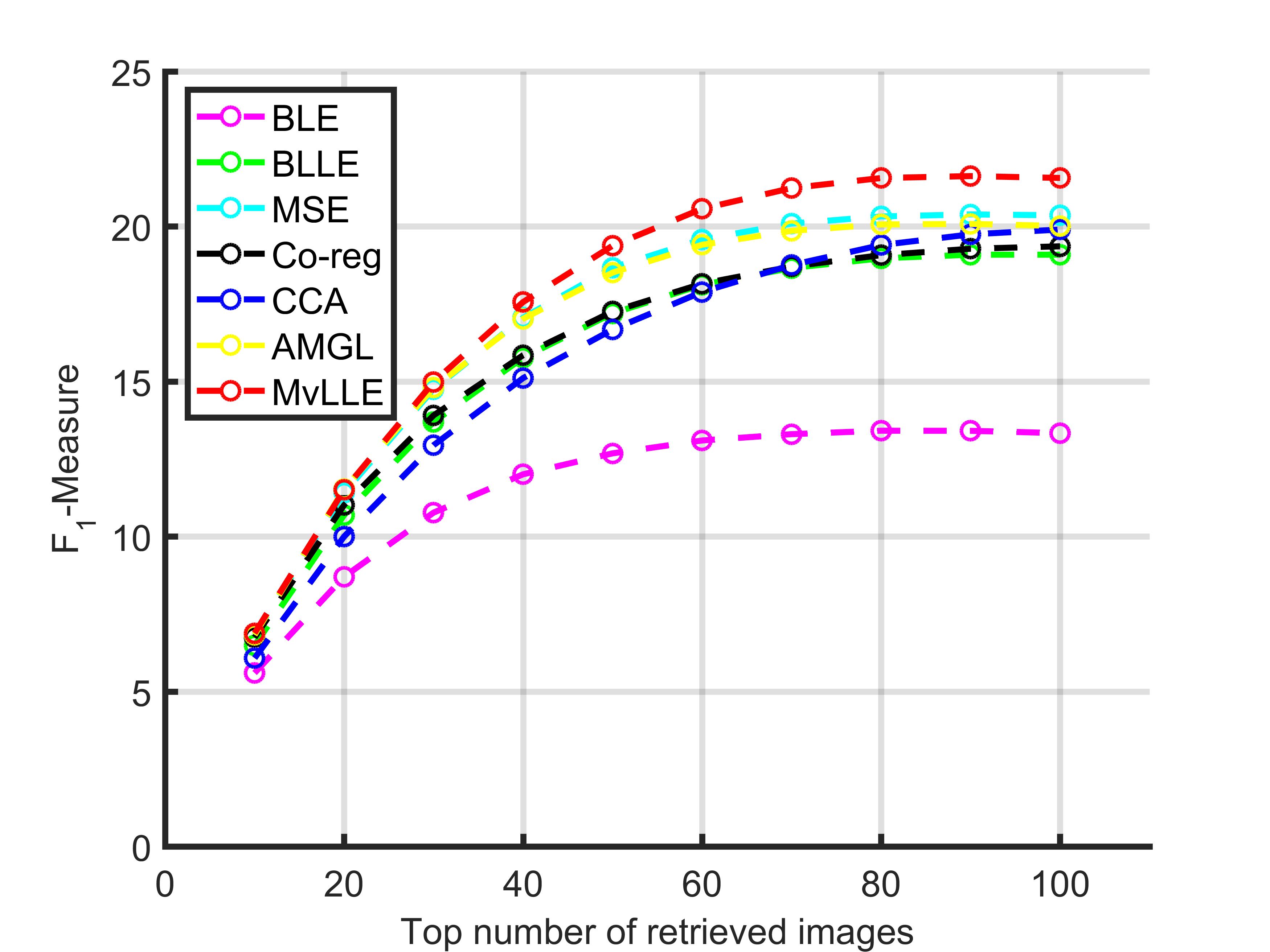}
}

\caption{The curves of precision, recall, PR, and $F_{1}$-Measure on Corel-1K dataset.}
\label{Corel1K}
\end{figure*}

In this section, we conduct two experiments on Holidays and Corel-1K datasets for image retrieval. For these two datasets, we both employ three image descriptors of MSD \cite{LIU2011image}, Gist \cite{douze2009evaluation}, and HOC \cite{yu2016A} to extract multi-view features for all images. All the methods are conducted to project all samples to the same dimensionality. In this part, the dimensions of the embedding obtained by all methods maintain 30 dimensions. Besides, $\mathop{l}_1$ distance is utilized to measure similarities between samples. At the aspect of the validation index, we choose several common indexes, including average precision rate (Precision), average recall rate (Recall), mean average precision (MAP), and $F_1$-Measure, to validate the performances for image retrieval. Actually, high Precision and Recall are required and $F_1$-Measure is put forward as the overall performance measurement. Then, we conducted this experiment on these two datasets repeatedly for twenty times. For Holidays dataset, we summarize these experiment results, including Precision, Recall, MAP, and $F_1$-Measure, on top 2 retrieval results in Table \ref{Holidays}. For Corel-1K dataset, we randomly select 10 images as query ones for each category. Afterward, the relation curves on validation indexes are drawn in Fig. \ref{Corel1K}.

\begin{table}[!htb]
\caption{The image retrieval accuracy on Holidays dataset.}
\label{Holidays}
\centering  %
\begin{tabular}{lllll}  
\hline
Methods & Precision (\%) & Recall (\%) & MAP (\%) & $F_1$-Measure \\
\hline
BLE & 72.92 & 56.16 & 86.46 & 31.73 \\
BLLE & 59.84 & 63.61 & 80.86 & 30.73 \\
MSE & 77.09 & 59.56 & 88.54 & 33.63 \\
Co-reg & 77.25 & 59.51 & 88.52 & 33.62 \\
CCA & 65.22 & 50.05 & 78.32 & 28.32 \\
AMGL & 68.09 & 51.92 & 84.01 & 29.46 \\
MvLLE & \textbf{79.13} & \textbf{61.14} & \textbf{89.56} & \textbf{34.49} \\
\hline
\end{tabular}
\end{table}

Through these experimental results in Table \ref{Holidays} and Fig. \ref{Corel1K}, it can be readily found that our proposed MvLLE achieves better performance than the other compared methods in most situations in the field of image retrieval. Our proposed method MvLLE could integrate compatible and complementary information from multiple views and obtain a better embedding from these views. Therefore, the results in Table \ref{Holidays} and Fig. \ref{Corel1K} could show that our framework can achieve good performance in the field of face recognition. Note that the performance of BLE is bad because of its unreasonable way to deal with multi-view features.

\subsection{Discussion}\label{discussion}
For the experiment results in Table \ref{3Source} and Table \ref{Cora} on text classification, we can find that MvLLE outperforms other comparing methods in most situations. Similar to the performance validation in text classification, our proposed MvLLE also obtain promising performance in face recognition tasks through the evaluations in Figs.\ref{yale_result}-\ref{orl_result}. As shown in Table \ref{Holidays} and Fig.\ref{Corel1K}, our method could be also utilized to execute the image retrieval task. From the above evaluations, it's readily seen that the representations obtained by our method could be more effective and suitable for multi-view features. Besides, other multi-view methods outperform the other single-view methods in most situations, which could show multi-view learning is a valuable research field indeed. Compared with BLLE, MvLLE could achieve significantly better performance by integrating
the complementary information among different views meanwhile preserving its intrinsic characteristic in each view. Note that the experimental results of our proposed MvLLE on six datasets are without fine-tuning, and usage of fine-tuning might further improve its performance. Besides, we find that MvLLE could converge within limited iterations in most experiments, which empirically indicates the fast speed of the convergence for our method.

\section{Conclusion}\label{conclusion}
In this paper, we propose a novel unified multi-view framework, named Graph Consensus Multi-view Learning Framework (GCMLF), to extend most of graph embedding works based on single view into the multi-view setting. , It encourages all views to learn with each other according to the complementarity among views and explores the heterogeneous graph structure in each view independently to preserve the diversity property among all views. Based on the sufficient theoretical analysis, we show that GCMLF is a more robust and flexible multi-view learning framework than those existing multi-view methods. Correspondingly, an algorithm based on alternating direction strategy is proposed to solve GCMLF, and the relative proof guarantees that it can converge to a local optimal solution. Furthermore, we provide one typical implement based on two heterogeneous graph embedding methods of LLE and LE, called Multi-view Locality Linear Embedding (MvLLE). Extensive experimental results demonstrate that the proposed MvLLE can effectively explore the diversity information and underlying complementary information of the given multi-view data, and outperforms its compared methods. With the rapid development of graph neural networks \cite{zhu2020cagnn, wu2020survey, zhu2021deep}, how to extend our framework into this domain is very meaningful yet full of challenges, and we will consider it in our future work.

\section*{Acknowledgements}
The authors would like to thank the anonymous reviewers for their insightful comments and suggestions to significantly improve the quality of this paper.
This work was supported by National Natural Science Foundation of PR China(61672130, 61972064) and LiaoNing Revitalization Talents Program(XLYC1806006).

\bibliographystyle{IEEEtran}
\bibliography{IEEEexample}

\begin{thebibliography}{10}
\providecommand{\url}[1]{#1}
\csname url@samestyle\endcsname
\providecommand{\newblock}{\relax}
\providecommand{\bibinfo}[2]{#2}
\providecommand{\BIBentrySTDinterwordspacing}{\spaceskip=0pt\relax}
\providecommand{\BIBentryALTinterwordstretchfactor}{4}
\providecommand{\BIBentryALTinterwordspacing}{\spaceskip=\fontdimen2\font plus
\BIBentryALTinterwordstretchfactor\fontdimen3\font minus
  \fontdimen4\font\relax}
\providecommand{\BIBforeignlanguage}[2]{{%
\expandafter\ifx\csname l@#1\endcsname\relax
\typeout{** WARNING: IEEEtran.bst: No hyphenation pattern has been}%
\typeout{** loaded for the language `#1'. Using the pattern for}%
\typeout{** the default language instead.}%
\else
\language=\csname l@#1\endcsname
\fi
#2}}
\providecommand{\BIBdecl}{\relax}
\BIBdecl

\bibitem{li2018survey}
Y.~Li, M.~Yang, and Z.~M. Zhang, ``A survey of multi-view representation
  learning,'' \emph{IEEE Transactions on Knowledge and Data Engineering},
  vol.~31, no.~10, pp. 1863--1883, 2018.

\bibitem{zhao2017multi}
J.~Zhao, X.~Xie, X.~Xu, and S.~Sun, ``Multi-view learning overview: Recent
  progress and new challenges,'' \emph{Information Fusion}, vol.~38, pp.
  43--54, 2017.

\bibitem{ojala2002multiresolution}
T.~Ojala, M.~Pietik{\"a}inen, and T.~M{\"a}enp{\"a}{\"a}, ``Multiresolution
  gray-scale and rotation invariant texture classification with local binary
  patterns,'' \emph{IEEE Transactions on Pattern Analysis and Machine
  Intelligence}, vol.~24, no.~7, pp. 971--987, 2002.

\bibitem{douze2009evaluation}
M.~Douze, H.~J{\'e}gou, H.~Sandhawalia, L.~Amsaleg, and C.~Schmid, ``Evaluation
  of gist descriptors for web-scale image search,'' in \emph{Proceedings of the
  ACM International Conference on Image and Video Retrieval}.\hskip 1em plus
  0.5em minus 0.4em\relax ACM, 2009, pp. 1--8.

\bibitem{gao2008image}
X.~Gao, B.~Xiao, D.~Tao, and X.~Li, ``Image categorization: Graph edit
  distance+ edge direction histogram,'' \emph{Pattern Recognition}, vol.~41,
  no.~10, pp. 3179--3191, 2008.

\bibitem{Amini2009Learning}
M.~R. Amini, N.~Usunier, and C.~Goutte, ``Learning from multiple partially
  observed views - an application to multilingual text categorization,'' in
  \emph{Advances in Neural Information Processing Systems}, 2009, pp. 28--36.

\bibitem{bisson2012co}
G.~Bisson and C.~Grimal, ``Co-clustering of multi-view datasets: a
  parallelizable approach,'' in \emph{International Conference on Data
  Mining}.\hskip 1em plus 0.5em minus 0.4em\relax IEEE, 2012, pp. 828--833.

\bibitem{kan2016multi}
M.~Kan, S.~Shan, and X.~Chen, ``Multi-view deep network for cross-view
  classification,'' in \emph{Proceedings of the IEEE Conference on Computer
  Vision and Pattern Recognition}, 2016, pp. 4847--4855.

\bibitem{zhang2019multi}
C.~Zhang, J.~Cheng, and Q.~Tian, ``Multi-view image classification with visual,
  semantic and view consistency,'' \emph{IEEE Transactions on Image
  Processing}, vol.~29, pp. 617--627, 2019.

\bibitem{Wang2019A}
H.~Wang, Y.~Yang, B.~Liu, and H.~Fujita, ``A study of graph-based system for
  multi-view clustering,'' \emph{Knowledge-Based Systems}, vol. 163, no. JAN.1,
  pp. 1009--1019, 2019.

\bibitem{Zheng2018Binary}
Z.~Zheng, L.~Li, F.~Shen, S.~H. Tao, and S.~Ling, ``Binary multi-view
  clustering,'' \emph{IEEE Transactions on Pattern Analysis and Machine
  Intelligence}, vol.~41, no.~7, pp. 1774--1782, 2018.

\bibitem{yang2018multi}
Y.~Yang and H.~Wang, ``Multi-view clustering: A survey,'' \emph{Big Data Mining
  and Analytics}, vol.~1, no.~2, pp. 83--107, 2018.

\bibitem{wang2019gmc}
H.~Wang, Y.~Yang, and B.~Liu, ``Gmc: Graph-based multi-view clustering,''
  \emph{IEEE Transactions on Knowledge and Data Engineering}, vol.~32, no.~6,
  pp. 1116--1129, 2019.

\bibitem{xia2010multiview}
T.~Xia, D.~Tao, T.~Mei, and Y.~Zhang, ``Multiview spectral embedding,''
  \emph{IEEE Transactions on Systems, Man, and Cybernetics, Part B
  (Cybernetics)}, vol.~40, no.~6, pp. 1438--1446, 2010.

\bibitem{Nie2017Auto}
F.~Nie, G.~Cai, J.~Li, and X.~Li, ``Auto-weighted multi-view learning for image
  clustering and semi-supervised classification,'' \emph{IEEE Transactions on
  Image Processing}, vol.~27, no.~3, pp. 1501--1511, 2017.

\bibitem{Huang2018Self}
S.~Huang, Z.~Kang, and Z.~Xu, ``Self-weighted multi-view clustering with soft
  capped norm,'' \emph{Knowledge-Based Systems}, vol. 158, no.~15, pp. 1--8,
  2018.

\bibitem{tian19a}
L.~Tian, F.~Nie, and X.~Li, ``A unified weight learning paradigm for multi-view
  learning,'' in \emph{Proceedings of Machine Learning Research},
  vol.~89.\hskip 1em plus 0.5em minus 0.4em\relax PMLR, 2019, pp. 2790--2800.

\bibitem{belkin2002laplacian}
M.~Belkin and P.~Niyogi, ``Laplacian eigenmaps and spectral techniques for
  embedding and clustering,'' in \emph{Advances in Neural Information
  Processing Systems}, 2002, pp. 585--591.

\bibitem{Wang2010A}
W.~Wang and Z.~H. Zhou, ``A new analysis of co-training,'' in
  \emph{International Conference on International Conference on Machine
  Learning}, 2010.

\bibitem{kumar2011co2}
A.~Kumar and H.~Daum{\'e}, ``A co-training approach for multi-view spectral
  clustering,'' in \emph{Proceedings of the 28th International Conference on
  Machine Learning}, 2011, pp. 393--400.

\bibitem{kumar2011co}
A.~Kumar, P.~Rai, and H.~Daume, ``Co-regularized multi-view spectral
  clustering,'' in \emph{Advances in Neural Information Processing Systems},
  2011, pp. 1413--1421.

\bibitem{Niu2019Multi}
X.~Niu, H.~Han, S.~Shan, and X.~Chen, ``Multi-label co-regularization for
  semi-supervised facial action unit recognition,'' in \emph{Advances in Neural
  Information Processing Systems}, 2019, pp. 909--919.

\bibitem{yan2006graph}
S.~Yan, D.~Xu, B.~Zhang, H.-J. Zhang, Q.~Yang, and S.~Lin, ``Graph embedding
  and extensions: A general framework for dimensionality reduction,''
  \emph{IEEE Transactions on Pattern Analysis and Machine Intelligence},
  vol.~29, no.~1, pp. 40--51, 2006.

\bibitem{wold1987principal}
S.~Wold, K.~Esbensen, and P.~Geladi, ``Principal component analysis,''
  \emph{Chemometrics and Intelligent Laboratory Systems}, vol.~2, no. 1-3, pp.
  37--52, 1987.

\bibitem{belhumeur1997eigenfaces}
P.~N. Belhumeur, J.~P. Hespanha, and D.~J. Kriegman, ``Eigenfaces vs.
  fisherfaces: Recognition using class specific linear projection,'' \emph{IEEE
  Transactions on Pattern Analysis and Machine Intelligence}, vol.~19, no.~7,
  pp. 711--720, 1997.

\bibitem{weinberger2006distance}
K.~Q. Weinberger, J.~Blitzer, and L.~K. Saul, ``Distance metric learning for
  large margin nearest neighbor classification,'' in \emph{Advances in Neural
  Information Processing Systems}, 2006, pp. 1473--1480.

\bibitem{scholkopf1997kernel}
B.~Sch{\"o}lkopf, A.~Smola, and K.-R. M{\"u}ller, ``Kernel principal component
  analysis,'' in \emph{International Conference on Artificial Neural
  Networks}.\hskip 1em plus 0.5em minus 0.4em\relax Springer, 1997, pp.
  583--588.

\bibitem{torresani2007large}
L.~Torresani and K.-c. Lee, ``Large margin component analysis,'' in
  \emph{Advances in Neural Information Processing Systems}, 2007, pp.
  1385--1392.

\bibitem{mika1999fisher}
S.~Mika, G.~Ratsch, J.~Weston, B.~Scholkopf, and K.-R. Mullers, ``Fisher
  discriminant analysis with kernels,'' in \emph{Neural networks for signal
  processing IX: Proceedings of the 1999 IEEE signal processing society
  workshop (cat. no. 98th8468)}.\hskip 1em plus 0.5em minus 0.4em\relax IEEE,
  1999, pp. 41--48.

\bibitem{zhu2020deep}
Y.~Zhu, Y.~Xu, F.~Yu, Q.~Liu, S.~Wu, and L.~Wang, ``Deep graph contrastive
  representation learning,'' in \emph{ICML Workshop on Graph Representation
  Learning and Beyond}, 2020.

\bibitem{zhu2021graph}
------, ``Graph contrastive learning with adaptive augmentation,'' in
  \emph{Proceedings of the Web Conference 2021}.\hskip 1em plus 0.5em minus
  0.4em\relax ACM Press, 2021, pp. 2069--2080.

\bibitem{roweis2000nonlinear}
S.~T. Roweis and L.~K. Saul, ``Nonlinear dimensionality reduction by locally
  linear embedding,'' \emph{Science}, vol. 290, no. 5500, pp. 2323--2326, 2000.

\bibitem{cao2013robust}
T.~Cao, V.~Jojic, S.~Modla, D.~Powell, K.~Czymmek, and M.~Niethammer, ``Robust
  multimodal dictionary learning,'' in \emph{International Conference on
  Medical Image Computing and Computer-Assisted Intervention}.\hskip 1em plus
  0.5em minus 0.4em\relax Springer, 2013, pp. 259--266.

\bibitem{liu2014multiview}
W.~Liu, D.~Tao, J.~Cheng, and Y.~Tang, ``Multiview hessian discriminative
  sparse coding for image annotation,'' \emph{Computer Vision and Image
  Understanding}, vol. 118, pp. 50--60, 2014.

\bibitem{Lin2011Multiple}
Y.~Y. Lin, T.~L. Liu, and C.~S. Fuh, ``Multiple kernel learning for
  dimensionality reduction,'' \emph{IEEE Transactions on Pattern Analysis and
  Machine Intelligence}, vol.~33, no.~6, pp. 1147--1160, 2011.

\bibitem{Nen2011Multiple}
M.~Nen, Alpayd, and N.~Ethem, ``Multiple kernel learning algorithms,''
  \emph{Journal of Machine Learning Research}, vol.~12, pp. 2211--2268, 2011.

\bibitem{Gu2017Multiple}
Y.~Gu, J.~Chanussot, X.~Jia, and J.~A. Benediktsson, ``Multiple kernel learning
  for hyperspectral image classification: A review,'' \emph{IEEE Transactions
  on Geoscience and Remote Sensing}, vol.~55, no.~11, pp. 6547--6565, 2017.

\bibitem{hardoon2004canonical}
D.~R. Hardoon, S.~Szedmak, and J.~Shawe-Taylor, ``Canonical correlation
  analysis: An overview with application to learning methods,'' \emph{Neural
  Computation}, vol.~16, no.~12, pp. 2639--2664, 2004.

\bibitem{gretton2005measuring}
A.~Gretton, O.~Bousquet, A.~Smola, and B.~Sch{\"o}lkopf, ``Measuring
  statistical dependence with hilbert-schmidt norms,'' in \emph{International
  conference on algorithmic learning theory}.\hskip 1em plus 0.5em minus
  0.4em\relax Springer, 2005, pp. 63--77.

\bibitem{rupnik2010multi}
J.~Rupnik and J.~Shawe-Taylor, ``Multi-view canonical correlation analysis,''
  in \emph{Conference on Data Mining and Data Warehouses}, 2010, pp. 1--4.

\bibitem{sharma2012generalized}
A.~Sharma, A.~Kumar, H.~Daume, and D.~W. Jacobs, ``Generalized multiview
  analysis: A discriminative latent space,'' in \emph{2012 IEEE Conference on
  Computer Vision and Pattern Recognition}.\hskip 1em plus 0.5em minus
  0.4em\relax IEEE, 2012, pp. 2160--2167.

\bibitem{kan2016multi-view}
M.~Kan, S.~Shan, H.~Zhang, S.~Lao, and X.~Chen, ``Multi-view discriminant
  analysis,'' \emph{IEEE Transactions on Pattern Analysis and Machine
  Intelligence}, vol.~38, no.~1, pp. 188--194, 2016.

\bibitem{Niu2014Iterative}
Niu, D., Dy, J, Jordan, and a, ``Iterative discovery of multiple
  alternativeclustering views,'' \emph{IEEE Transactions on Pattern Analysis
  and Machine Intelligence}, vol.~36, no.~7, pp. 1340--1353, 2014.

\bibitem{Cao2015Diversity}
X.~Cao, C.~Zhang, H.~Fu, S.~Liu, and H.~Zhang, ``Diversity-induced multi-view
  subspace clustering,'' in \emph{Computer Vision and Pattern Recognition},
  2015, pp. 586--594.

\bibitem{Zhang2016Flexible}
C.~Zhang, H.~Fu, Q.~Hu, P.~Zhu, and X.~Cao, ``Flexible multi-view
  dimensionality co-reduction,'' \emph{IEEE Transactions on Image Processing},
  vol.~26, no.~2, pp. 648--659, 2016.

\bibitem{liu2013robust}
G.~Liu, Z.~Lin, S.~Yan, J.~Sun, Y.~Yu, and Y.~Ma, ``Robust recovery of subspace
  structures by low-rank representation,'' \emph{IEEE Transactions on Pattern
  Analysis and Machine Intelligence}, vol.~35, no.~1, pp. 171--184, 2013.

\bibitem{rudin1964principles}
W.~Rudin \emph{et~al.}, \emph{Principles of mathematical analysis}.\hskip 1em
  plus 0.5em minus 0.4em\relax McGraw-hill New York, 1964, vol.~3.

\bibitem{belkin2003laplacian}
M.~Belkin and P.~Niyogi, ``Laplacian eigenmaps for dimensionality reduction and
  data representation,'' \emph{Neural Computation}, vol.~15, no.~6, pp.
  1373--1396, 2003.

\bibitem{LIU2011image}
L.~Guang-Hai, L.~Zuo-Yong, Z.~Lei, and X.~Yong, ``Image retrieval based on
  micro-structure descriptor,'' \emph{Pattern Recognition}, vol.~44, no.~9, pp.
  2123 -- 2133, 2011.

\bibitem{yu2016A}
L.~Yu, L.~Feng, C.~Chen, T.~Qiu, and J.~Wu, ``A novel multi-feature
  representation of images for heterogeneous iots,'' \emph{IEEE Access},
  vol.~4, no.~99, pp. 6204--6215, 2016.

\bibitem{zhu2020cagnn}
Y.~Zhu, Y.~Xu, F.~Yu, S.~Wu, and L.~Wang, ``Cagnn: Cluster-aware graph neural
  networks for unsupervised graph representation learning,'' \emph{arXiv
  preprint arXiv:2009.01674}, 2020.

\bibitem{wu2020survey}
Z.~Wu, S.~Pan, F.~Chen, G.~Long, C.~Zhang, and P.~S. Yu, ``A comprehensive
  survey on graph neural networks,'' \emph{IEEE Transactions on Neural Networks
  and Learning Systems}, pp. 1--21, 2020.

\bibitem{zhu2021deep}
Y.~Zhu, W.~Xu, J.~Zhang, Q.~Liu, S.~Wu, and L.~Wang, ``Deep graph structure
  learning for robust representations: A survey,'' \emph{arXiv preprint
  arXiv:2103.03036}, 2021.

\end{thebibliography}

\vspace{12pt}

\begin{IEEEbiography}[{\includegraphics[width=1in,height=1.25in,clip,keepaspectratio]{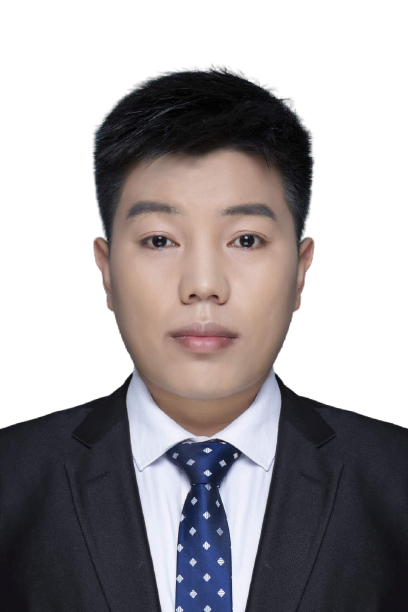}}]{Xiangzhu Meng} received his BS degree from Anhui University, in 2015. Now he is working towards the PHD degree in School of Computer Science and Technology, Dalian University of Technology, China. He has authored and co-authored some papers in some famous journals, including Knowledge-Based Systems, Engineering Applications of Artificial Intelligence, Neurocomputing, etc. Furthermore, he serves as a reviewer for ACM Transactions on Multimedia Computing Communications and Applications. His research interests include multi-view learning, deep learning, data mining and computing vision.
\end{IEEEbiography}

\begin{IEEEbiography}[{\includegraphics[width=1in,height=1.25in,clip,keepaspectratio]{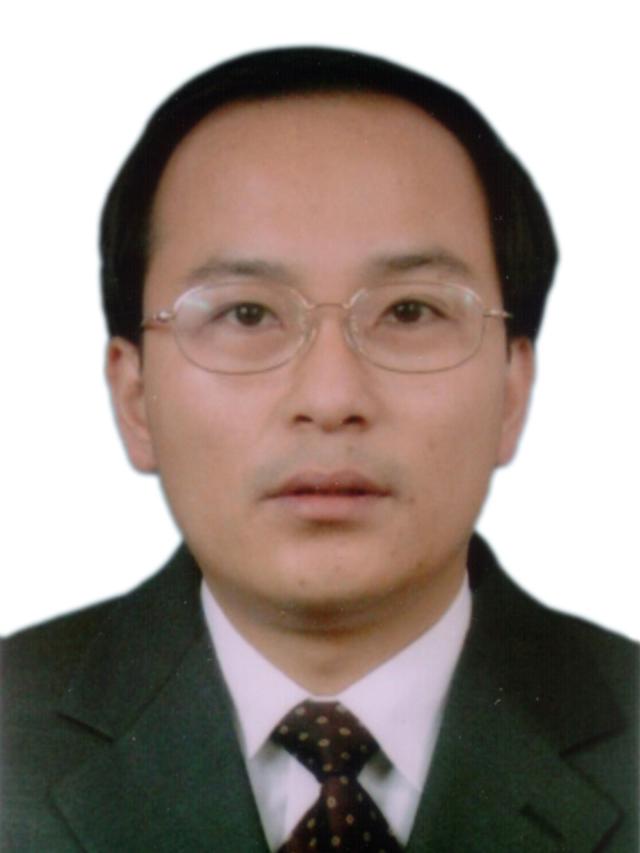}}]{Lin Feng}
received the BS degree in electronic technology from Dalian University of Technology, China, in 1992, the MS degree in power engineering from Dalian University of Technology, China, in 1995, and the PhD degree in mechanical design and theory from Dalian University of Technology, China, in 2004. He is currently a professor and doctoral supervisor in the School of Innovation Experiment, Dalian University of Technology, China. His research interests include intelligent image processing, robotics, data mining, and embedded systems.
\end{IEEEbiography}

\begin{IEEEbiography}[{\includegraphics[width=1in,height=1.25in,clip,keepaspectratio]{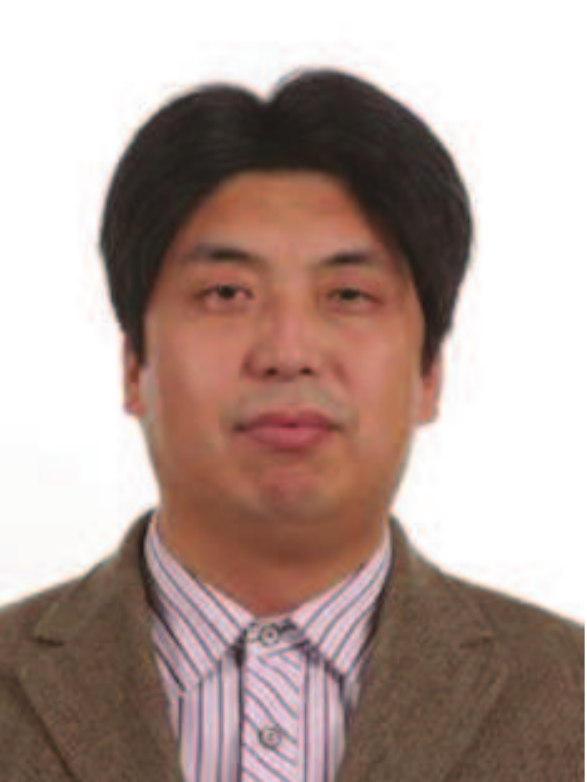}}]{Chonghui Guo}
received the B.S. degree in mathematics from Liaoning University, in 1995, the M.S. degree in operational research and control theory, and the Ph.D. degree in Institute of Systems Engineering from the Dalian University of Technology, in 2002. He is a professor of the Institute of Systems Engineering, Dalian University of Technology. He was a postdoctoral research fellow in the Department of Computer Science, Tsinghua University. His interests include data mining and knowledge discovery.
\end{IEEEbiography}

\end{document}